\documentclass[lettersize,journal]{IEEEtran}
\usepackage{amsmath,amsfonts}
\usepackage{algorithmic}
\usepackage{algorithm}
\usepackage{array}
\usepackage[caption=false,font=normalsize,labelfont=sf,textfont=sf]{subfig}
\usepackage{textcomp}
\usepackage{stfloats}
\usepackage{steinmetz}
\usepackage{url}
\usepackage{float} 
\usepackage{color}
\usepackage{verbatim}
\usepackage{graphicx}
\usepackage{cite}
\usepackage{multirow}  
\usepackage{tabularray}
\usepackage[
colorlinks,
linkcolor=red,
anchorcolor=green,
filecolor=red,
urlcolor=red,
citecolor=green]{hyperref}
\begin{document}

\title{No-Reference Point Cloud Quality Assessment via Graph Convolutional Network}

\author{Wu Chen, 
        Qiuping Jiang,~\IEEEmembership{Senior Member,~IEEE},
        Wei Zhou,~\IEEEmembership{Senior Member,~IEEE},
        Feng Shao,~\IEEEmembership{Senior Member,~IEEE},
        Guangtao Zhai,~\IEEEmembership{Senior Member,~IEEE},
        Weisi Lin,~\IEEEmembership{Fellow,~IEEE}

\thanks{
       This work was supported in part by the Natural Science Foundation of China (62271277, 62301323), in part by the Natural Science Foundation of Zhejiang (LR22F020002), in part by the “Leading Goose” R\&D Program of Zhejiang Province (2024C01129), in part by the Key R\&D Program of Ningbo (2024Z292), and in part by the Shenzhen Natural Science Foundation (20231128191435002). (\textit{Corresponding author: Qiuping~Jiang})

        W. Chen, Q. Jiang, and F. Shao are with the School of Information Science and 
        Engineering, Ningbo University, Ningbo 315211, China (e-mail: 2212100375@nbu.edu.cn; jiangqiuping@nbu.edu.cn; shaofeng@nbu.edu.cn).

        W. Zhou is with the School of Computer Science and Informatics, Cardiff University, Cardiff, United Kingdom (e-mail: zhouw26@cardiff.ac.uk)
			
		G. Zhai is with the Institute of Image Communication and Information Processing, Shanghai Jiao Tong University, Shanghai 200240, China (zhaiguangtao@sjtu.edu.cn)

        W. Lin is with the School of Computer Science and Engineering, Nanyang Technological University, Singapore (wslin@ntu.edu.sg)
        
        %Corresponding author: Qiuping Jiang
        }}

\markboth{}%
{Shell \MakeLowercase{\textit{et al.}}: A Sample Article Using IEEEtran.cls for IEEE Journals}

\maketitle

\begin{abstract} 
Three-dimensional (3D) point cloud, as an emerging visual media format, is increasingly favored by consumers as it can provide more realistic visual information than two-dimensional (2D) data. Similar to 2D plane images and videos, point clouds inevitably suffer from quality degradation and information loss through multimedia communication systems. Therefore, automatic point cloud quality assessment (PCQA) is of critical importance. In this work, we propose a novel no-reference PCQA method by using a graph convolutional network (GCN) to characterize the mutual dependencies of multi-view 2D projected image contents. The proposed GCN-based PCQA (GC-PCQA) method contains three modules, i.e., multi-view projection, graph construction, and GCN-based quality prediction. First, multi-view projection is performed on the test point cloud to obtain a set of horizontally and vertically projected images. Then, a perception-consistent graph is constructed based on the spatial relations among different projected images. Finally, reasoning on the constructed graph is performed by GCN to characterize the mutual dependencies and interactions between different projected images, and aggregate feature information of multi-view projected images for final quality prediction. Experimental results on two publicly available benchmark databases show that our proposed GC-PCQA can achieve superior performance than state-of-the-art quality assessment metrics. 
The code will be available at: \url{https://github.com/chenwuwq/GC-PCQA}.
\end{abstract}


% 引言
\section{Introduction}
\IEEEPARstart{I}{n} recent years, the development of three-dimensional (3D) visual information acquisition technology makes point clouds easier to obtain and gradually becomes a popular type of visual data. A 3D Point cloud is mainly used to describe a complete 3D scene or object, including geometric attributes (position of each point in 3D space), color attributes (RGB attributes of each point), and others (normal vector, opacity, reflectivity, time, etc.)\cite{liu2020model}. Point clouds have been widely studied and used in a wide range of application scenarios such as 3D reconstruction\cite{huang2020pf,chen2020vis}, classification and segmentation\cite{PointNet,PointNet++}, facial expression representation\cite{demisse2018def}, autonomous driving\cite{javaheri2020point,cui2021deep}, and virtual reality\cite{alexiou2020pointxr}, etc. Although point cloud can realistically record 3D objects through a large set of points, it also consumes a lot of memory, and it is difficult to achieve data transmission under limited network bandwidth\cite{schwarz2018mpeg,gu20203d}. This new and effective data representation presents a challenge to the current hardware storage and network transmission. 
Therefore, in order to achieve efficient storage and transmission, compression of point clouds is necessary\cite{vpcc,gpcc,liu2020coarse,xu2021point}. However, point cloud compression may introduce artifacts, resulting in the degradation of point cloud visual quality. Point cloud visual quality is an important way to compare the performance of various point cloud processing algorithms. Effective point cloud quality assessment (PCQA) methods can not only help people evaluate the distortion degree of point clouds and the performance of compression algorithms but also be beneficial to optimize the visual quality of distorted point clouds. Thus, how to accurately assess the perceptual quality of point clouds has become a critical issue.

Similar to image quality assessment (IQA), PCQA can also be divided into subjective and objective methods. The subjective method is mainly based on the perception of the human visual system (HVS). It is difficult to be widely applied because this kind of assessment requires a large number of participants to ensure the rationality and accuracy of the assessment results in a statistical sense. 
Currently, the results obtained from subjective assessment experiments are generally served as the ground-truth data for benchmarking different objective methods \cite{perry2020quality}. According to the participation of original point clouds, objective PCQA methods can have three categories: full reference (FR), reduced reference (RR), and no reference (NR). Since the original point clouds are not always available, NR-PCQA methods that do not rely on any original information as a reference are more suitable in practical applications. 

The traditional NR-PCQA methods \cite{hua2021bqe, zhang2022no} generally predict the quality score by extracting quality-aware features based on the analysis of point cloud attributes such as geometry and color. Recently, the great success of deep learning in the field of NR-IQA has promoted the development of deep learning-based NR-PCQA metrics \cite{chetouani2021deep, liu2023point, tao2021point, liu2021pqa, tu2022v, yang2022no}. The common practice of these deep NR-PCQA metrics is to directly apply the ordinary convolution operation on the point cloud for automatic feature learning in a data-driven manner. Nonetheless, point cloud is a typical kind of non-Euclidean data which is sparsely distributed over the 3D space, and a large number of useless pixels are also involved with pixel-by-pixel convolution, thus resulting in a huge waste of resources and inefficient data processing. In order to solve this problem, some related works try to represent non-Euclidean data with the graph  which includes node information and complex adjacency relations between nodes. With the graph-based non-Euclidean data as input, the current works then introduce to use graph convolutional network (GCN) rather than traditional convolutional neural network (CNN) for more effective feature representation learning \cite{wu2020compre}. For instance, Thomas et al.\cite{kipf2016semi} proposed to convert non-Euclidean data into a graph based on which the GCN is used to realize graph feature extraction. As a typical kind of non-Euclidean data, GCN has also been applied to many point cloud-based vision tasks, such as point cloud classification\cite{zhang2018graph, li2021structural}, point cloud segmentation\cite{xu2019msgcnn}, point cloud data analysis\cite{hansen2018multi}, action recognition\cite{si2019action}, etc. Moreover, it has also been applied to infer the perceptual quality of various multimedia data, e.g., traditional 2D images \cite{sun2022graphiqa,huang2022explainable}, 360-degree images \cite{xu2020blind,fu2022adaptive}, and meshes \cite{abouelaziz2021learning,el2022learning}. 

\begin{figure}[t]
	\centering
	\includegraphics[width=\linewidth]{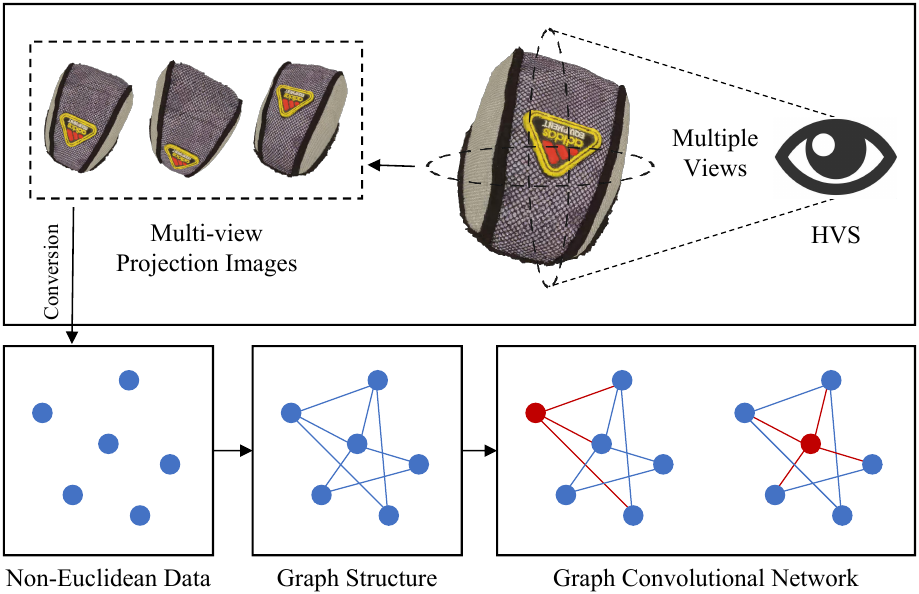}
	\caption{We simulate the perceptual process of HVS to perform multi-view projection on the 3D point cloud and build the graph based on the projected images. The red node in graph convolution represents the central node of the current convolution process, and the red line represents the adjacency relationship. The central node will constantly exchange information with neighboring nodes to aggregate feature information from neighbors.}
	\label{fig_1}
\end{figure}

Due to the strong capability of GCN in handling non-Euclidean data including 3D point cloud, this paper presents a novel GCN-based NR-PCQA method (GC-PCQA). One of the most critical issues is to effectively create a graph of the point cloud so that the GCN can be applied for feature learning. Since the goal of PCQA is to predict the quality of the test point cloud consistent with human perception, how to construct a highly perception-consistent graph of point clouds is the key to its success. It is known that the HVS reconstructs 3D objects in their mind based on multiple two-dimensional (2D) plane images observed from different viewpoints. In order to imitate the process of the HVS to perceive 3D objects, it is natural to perform multi-view projection on the point cloud to obtain a set of projected images with each corresponding to a specific viewpoint. Although these projected images are independent individuals, there is a certain extent of correlation between each other. Therefore, we regard all projected images as a set of non-Euclidean data and then establish a graph according to the dependencies between each individual projected image. Finally, GCN is applied to realize feature extraction from the constructed graph for quality prediction. The entire process is simply illustrated in Fig. \ref{fig_1}. Experimental results demonstrate that our proposed GC-PCQA method outperforms state-of-the-art reference and non-reference PCQA methods on two public PCQA databases. Overall, the main contributions of this paper are as follows:
\begin{enumerate}{}{}
	\item{We perform multi-view projection on the point cloud to obtain a set of projected images based on which a highly perception-consistent graph is constructed to model the mutual dependencies of multi-view projected images. The graph nodes are defined with the projected images and connected by spatial relations between each other.}
	\item{We perform GCN on the proposed graph to characterize the interactions between different projected images and aggregate the feature information of multi-view projected images for final quality prediction. The ablation study validates the effectiveness of the GCN architecture and the source code is available for public research usage.}
	\item{We fuse the horizontally and vertically projected image features extracted by two GCNs that do not share weights to boost the performance. Experimental results show that the proposed GC-PCQA can predict subjective scores more accurately than the existing state-of-the-art PCQA metrics.}
\end{enumerate}

The rest of this paper is organized as follows. In Section II, we introduce the related works. In Section III, we illustrate the proposed GC-PCQA with technical details. We conduct experiments and analyze the results in Section IV, and finally draw conclusions in Section V.

% 相关工作
\section{Related Work}
PCQA metrics have developed rapidly and can be mainly divided into PC-based metrics and projection-based metrics. PC-based metrics evaluate the quality score through the characteristic information of each point in the point cloud. While projection-based metrics use the projected images of the point cloud instead of the point cloud itself.

\subsection{PC-based Metrics}
As one of the important evaluation methods, the FR method has been widely investigated in PC-based metrics. The initial methods calculate quality scores based on geometric information of point clouds, such as $PSNR_{MSE,p2po}$ and $PSNR_{HF,p2po}$\cite{mekuria2016eval}, $PSNR_{MSE,p2pl}$ and $PSNR_{HF,p2pl}$\cite{tian2017geometric}. Among them, the point-to-point methods (p2point) compute the $L_2$ norm of the nearest point pair as the distortion measure of the point, while point-to-plane methods (p2plane) increase the normal vector of the plane. Besides, Alexiou et al.\cite{alexiou2018point} captured the distortion point cloud degradation through the angular similarity between the corresponding points. 
Javaheri et al.\cite{javaheri2020gen} used the generalized Hausdorff distance for PCQA. In addition to geometric properties, the color properties of point clouds can also be used as one of the important features for quality assessment. $PSNR_{Y}$\cite{mekuria2017per} evaluates texture distortion of colored point clouds based on point-to-point color components. 
Viola et al.\cite{viola2020color} used global color statistics, such as color histograms and correlograms, to evaluate the degree of distortion of a point cloud. 
On the basis of PC-MSDM\cite{meynet2019pc}, Meynet et al.\cite{meynet2020pcqm} proposed a linear model PCQM based on curvature and color attributes to predict 3D point cloud visual quality. Inspired by the idea of similarity, Alexiou et al.\cite{alexiou2020to} used the structural similarity index based on geometric and color features for evaluation. 
Diniz et al.\cite{diniz2020to,diniz2020multi,diniz2020local} extracted statistical information of point clouds based on local binary pattern descriptors and local luminance pattern descriptors, which are assessed by distance metrics. 
Yang et al.\cite{yang2020infer} proposed to construct the local graph representation of the reference point cloud and the distorted point cloud respectively with the key point as the center and calculate the similarity feature by extracting three color gradient moments between the central key point and all other points, so as to estimate the quality score of the distorted point cloud. They also believed that point clouds have potential energy, and used multiscale potential energy discrepancy (MPED)\cite{yang2022mped} to quantify point cloud distortion. Furthermore, Viola et al.\cite{viola2020red} extracted part of the geometric, color, and normal features from the point clouds, and then evaluated the distorted point cloud by finding the best combination of features through a linear optimization algorithm. Liu et al.\cite{liu2021reduced} proposed an analytical model with only three parameters to accurately predict the MOS of V-PCC compressed point clouds from geometric and color features. All of these methods use reference point clouds, and despite the advanced performance achieved, these methods may not be useful in practical applications. Therefore, it is meaningful to research NR methods to overcome the problem of missing reference point clouds. 
Zhang et al.\cite{zhang2022no} used 3D natural scene statistics (3D-NSS) and entropy to extract geometric and color features related to quality, and then predicted quality scores through a support vector regression (SVR) model. With structure-guided resampling, Zhou et al. \cite{zhou2022blind} estimated point cloud quality based on geometry density, color naturalness, and angular consistency. The success of deep learning in various research fields has prompted researchers to introduce it into PCQA. Chetouani et al.\cite{chetouani2021deep} used deep neural networks (DNNs) to learn the mapping of low-level features such as geometric distance, local curvature, and luminance values to quality scores. Liu et al.\cite{liu2023point} constructed a large-scale PCQA dataset named LS-PCQA, which contains more than 22,000 distortion samples, and then proposed an NR metric based on sparse CNN.

\subsection{Projection-based Metrics}
In addition to the above methods that directly use point clouds for quality evaluation, projection-based metrics also play an important role in PCQA. The projection-based PCQA metrics project the point cloud from 3D space to 2D plane, so as to transform PCQA into IQA which has been relatively mature. Therefore, the existing IQA methods can be directly used to evaluate the quality of 2D projection images, such as PSNR\cite{wolf2009ref}, SSIM\cite{wang2004image}, MS-SSIM\cite{wang2003multi}, IW-SSIM\cite{wang2010info}, VIFP\cite{sheikh2006image}, etc. 
Freitas et al.\cite{freitas2022point} used a multi-scale rotation invariant texture descriptor called Dominant Rotated Local Binary Pattern (DRLBP) to extract statistical features from these texture maps and calculate texture similarity. Finally, texture features and similarity features were fused to predict the visual quality of point clouds. 
Hua et al.\cite{hua2021bqe} proposed a blind quality evaluator of colored point cloud based on visual perception, which reduces the influence of visual masking effect by projecting the point cloud onto a plane to extract geometric, color and joint features. 
Tao et al.\cite{tao2021point} projected the color point cloud in 3D space into a 2D color and geometric projection map, and then weighted the quality scores of local blocks in the map based on a multi-scale feature fusion network. 
Liu et al.\cite{liu2021pqa} projected the six planes of the 3D point cloud, and then extracted multi-view features through a DNN to classify the distortion types of the point cloud. Finally, the final quality score was obtained by multiplying the probability vector and the quality vector. Tu et al.\cite{tu2022v} designed a two-stream CNN to extract the features of texture projection maps and geometric projection maps. Yang et al.\cite{yang2022no} used natural images as the source domain and point clouds as the target domain, and predicted point cloud quality through unsupervised adversarial domain adaptation.

Based on the above statements, PC-based metrics and projection-based metrics have achieved certain results. However, most of the existing projection-based PCQA metrics are evaluated based on six projection planes, which do not take into account the quality perception of HVS for 3D point clouds from multiple views, and do not make full use of the correlation between different projected images for modeling. Thus, we propose a novel non-reference PCQA method by using GCN to characterize the mutual dependencies of multi-view 2D projected image contents.

\begin{figure*}[t]
	\centering
	\includegraphics[width=\linewidth]{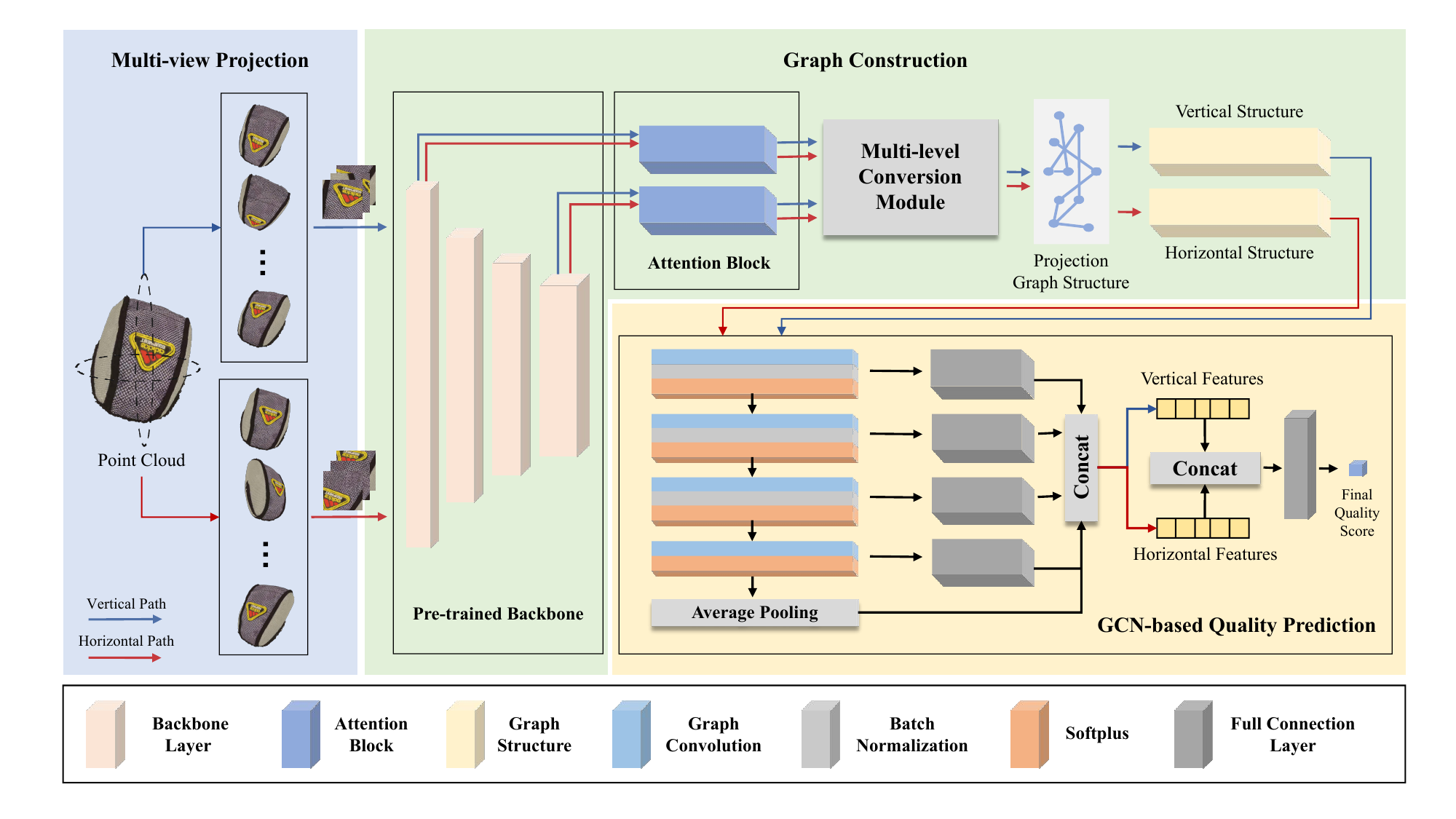}
	\caption{Framework of the proposed GC-PCQA method. It is mainly composed of three parts: multi-view projection, graph construction, and GCN-based quality prediction. Firstly, multi-view projection is performed on the point cloud to obtain a set of horizontally and vertically projected 2D images. All these projected images are fed into a pre-trained backbone and an attention block for attentive feature extraction. Secondly, a multi-level fusion of attentive feature maps is carried out through the multi-level conversion module, and graph construction is performed according to spatial relations among different projected images. Thirdly, reasoning on the constructed graph is performed by GCN to model the mutual dependencies between nodes and generate more effective feature representations. Finally, multi-level feature fusion is carried out on the feature representations obtained by two GCNs to predict the final quality score.}
	\label{fig_2}
\end{figure*}

\section{Proposed Method}
We first give a brief overview of the proposed GC-PCQA method. Then, the details of each module in our GC-PCQA method will be illustrated. Finally, we describe how the network is trained.

\subsection{Overview}
The framework of our proposed GC-PCQA method is shown in Fig. \ref{fig_2}. It is mainly composed of three parts: multi-view projection, graph construction, and GCN-based quality prediction. Firstly, considering the behavior of the HVS when observing 3D point clouds, multi-view projection is performed on the point cloud to obtain a set of horizontally and vertically projected 2D images. The horizontally (vertically) projected 2D image set covers the visual contents that can be perceived within the horizontal (vertical) visual field by an observer. All these projected images are fed into a pre-trained backbone and an attention block for attentive feature extraction. Secondly, a multi-level fusion of attentive feature maps is carried out through the multi-level conversion module, and graph construction is performed according to spatial relations among different projected images. Thirdly, reasoning on the constructed graph is performed by GCN to model the mutual dependencies between nodes and generate more effective feature representations. Finally, multi-level feature fusion is carried out on the feature representations obtained by two GCNs (corresponding to the horizontal visual field and the vertical visual field, respectively) to predict the final quality score.

\subsection{Multi-view Projection}
Point cloud consists of huge point sets, which is mainly used to describe 3D objects in detail, resulting in a large volume of point cloud and it is difficult to directly input point cloud data into the network. In order to reduce the cost of processing large-scale point clouds, many point cloud resampling strategies\cite{chen2017contour,chen2017fast,qi2019feature} and point cloud projection methods\cite{tao2021point,liu2021pqa,tu2022v,yang2022no} have been proposed to simplify point cloud data. Since deep learning is very effective in the field of image processing, we choose to use the multi-view projection method to convert the point cloud into an image, so as to take advantage of deep learning for PCQA.

The visual field of the human eye is divided into horizontal visual field and vertical visual field. The range of view in different directions is limited, i.e., the horizontal view limit is approximately 190 degrees while the vertical view limit is 135 degrees. When human eyes observe 3D objects such as 3D point clouds, it is difficult to directly observe all contents within 360 degrees\cite{xu2020blind}. In general, human reconstructs 3D objects in the brain by observing from different viewpoints, so as to have a clearer perception of 3D objects. 
In order to imitate the perceptual process of the HVS, we perform a multi-view projection operation on the point cloud based on rotation stride (RS).
Just as the human eye perceives 3D objects, we rotate and project the point cloud onto the 2D space in both horizontal and vertical directions. Finally, for each direction, we obtain a multi-view projected image group $\bf{P}$ containing $N$ projected images:
\begin{equation}
	\label{deqn_ex1}
	{\bf{P}} = [{\bf{Y}}_1,{\bf{Y}}_2,{\bf{Y}}_3,\cdots,{\bf{Y}}_N] \in \mathbb{R} ^ {N\times 3\times H\times W},
\end{equation}
where ${\bf{Y}}_{i}$ represents the $i$-th projected image, and $N$ is the total number of projected images. $H$ and $W$ indicate the height and width of each individual projected image, respectively. We use ${\bf{P}}_{H}$ to denote the horizontally projected image group and ${\bf{P}}_{V}$ to denote the vertically projected image group.

\subsection{Graph Construction}
The graph construction module aims to construct a perception-consistent graph based on the features extracted from multi-view projected images. It is mainly composed of a pre-trained backbone network, an attention block, and a multi-level conversion module.
\subsubsection{Feature Extraction}
We adopt the pre-trained ResNet101\cite{he2016deep} as the backbone for feature extraction. The input image size of the backbone network is 224$\times$224. 
However, due to the presence of a large amount of quality-unrelated white regions in the projected images, there may be adverse effects or an inability to learn quality-related features. To address this, an additional image pre-processing step is performed.
The projected image is cropped to effectively remove the non-informative white background regions. Subsequently, the cropped image is resized to 224$\times$224 to create an informative image patch, which is then used as input to the network.

\subsubsection{Attention Block}
\begin{figure}[t]
	\centering
	\includegraphics[width=\linewidth]{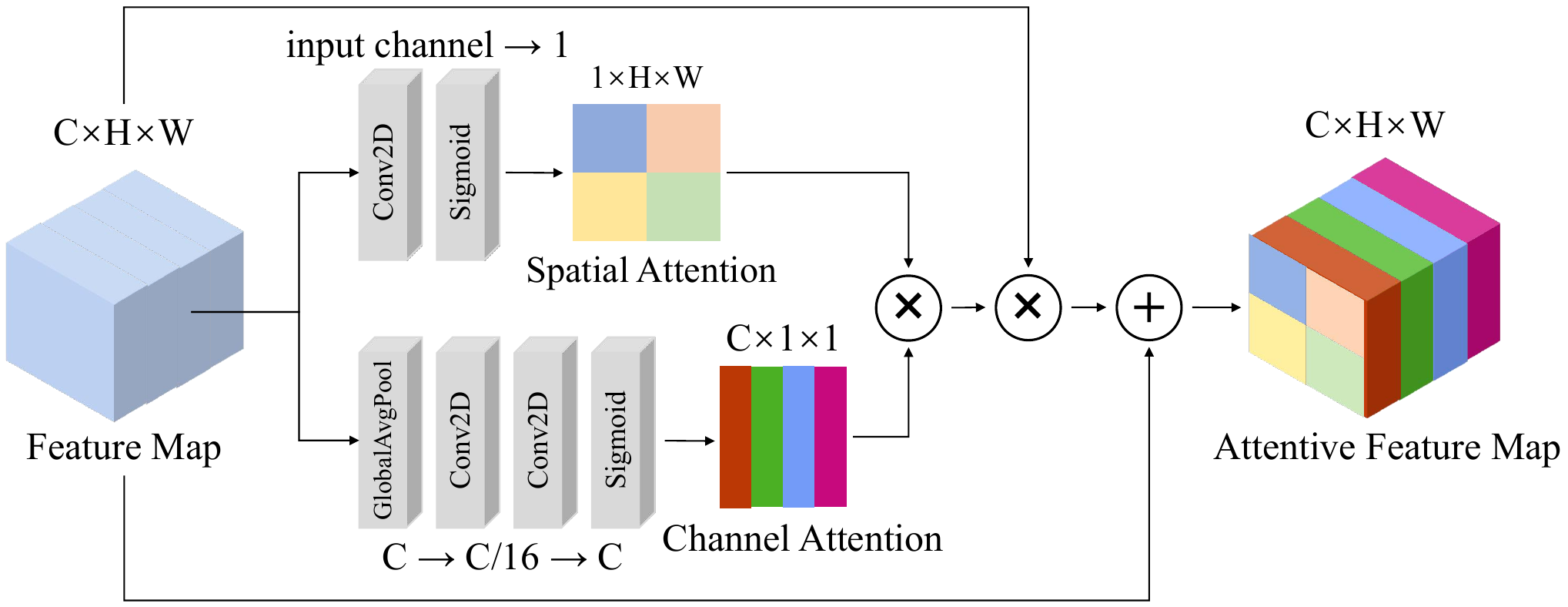}
	\caption{The attention block is used on the feature map extracted by the pre-trained backbone to obtain the attentive feature map. The upper part extracts spatial attention for the input feature map, and the lower part extracts channel attention. $H$: image height, $W$: image width, $C$: image channel.}
	\label{fig_3}
\end{figure}

The attention mechanism comes from the research on human vision and draws lessons from the attention thinking of the human vision, which can make the feature extractor focus more on those significant areas of the target while suppressing the most unimportant information to improve the performance of DNNs. At present, a variety of attention modules have been proposed such as SE attention\cite{hu2018se}, CBAM attention\cite{woo2018cbam}, scSE attention\cite{roy2018concurrent}, etc. 
Therefore, in order to further improve the feature representation capability of the network, we devise an attention block to impose appropriate attention weights to the feature maps obtained by the pre-trained backbone. The structure of our devised attention block is shown in Fig. \ref{fig_3}. 

We first introduce the upper flow, i.e., the spatial attention. The input feature maps from the pre-trained backbone can be represented as ${\bf{F}}^{(i)} \in\mathbb{R}^{\emph{C}\times H\times W}$, where $i$ represents the output of the $i^{th}$ layer of the backbone. Firstly, 2D convolution with a convolution kernel size of 1 is used to reduce the number of channels to 1. Then, a sigmoid activation function is applied to map the range of feature values into [0,1]. Finally, the spatial attention map ${\bf{F}}_{S}\in\mathbb{R}^{H\times W}$ is expressed as follows:
\begin{equation}
	\label{deqn_ex2}
    {\bf{F}}_{S}=\delta(\phi({\bf{F}}))=\left[\begin{array}{cccc}
                                        \delta(S_{11}) & \delta(S_{12}) & \cdots & \delta(S_{1W})\\
                                        \delta(S_{21}) & \delta(S_{22}) & \cdots & \delta(S_{2W})\\
                                        \vdots & \vdots & \vdots & \vdots\\ 
                                        \delta(S_{H1}) & \delta(S_{H2}) & \cdots & \delta(S_{HW})\\
                                        \end{array}\right],
\end{equation}
where $\phi$ represents the 2D convolution with a convolution kernel size of 1, $\delta(\cdot)$ denotes the sigmoid function, and $S_{ij}(i\in\{1,2,…,H\},j\in\{1,2,…,W\})$ indicates the relative importance of the eigenvalues at position $(i,j)$. 

For the lower flow, global average pooling is first applied on the input feature maps ${\bf{F}} \in\mathbb{R}^{\emph{C}\times H\times W}$. Then, the channel dimension of the feature maps is reduced and then increased by two 2D convolution layers with a convolution kernel size of 1. The final channel attention map ${\bf{F}}_{C} \in\mathbb{R}^{C\times 1\times 1}$ is generated by attaching a sigmoid activation function $\delta(\cdot)$ in the end, which can be expressed as follows:
\begin{equation}
	\label{deqn_ex3}
	{\bf{F}}_{C}=\delta(\phi(avg({\bf{F}}))))=[\delta(U_{1}),\delta(U_{1}),\cdots,\delta(U_{C})],
\end{equation}
where $U_{i}(i\in\{1,2,…,C\})$ denotes the relative importance of the $i^{th}$ channel among all channels, $avg$ denotes the global average pooling. After obtaining two attentions, the spatial attention map and the channel attention map are multiplied to obtain a mixed attention map ${\bf{F}}_{SC}\in\mathbb{R}^{C\times H\times W}$, which makes the information important in both space and channel dimensions more prominent and will encourage the network to learn more meaningful features. Then, the skip connection is used to multiply the original feature map and the mixed attention map pixel-by-pixel to complete information calibration. Finally, the residual connection is used to alleviate the gradient disappearance problem caused by increasing depth in the DNN. Mathematically, the final attentive feature map $\hat{{\bf{F}}}$ is generated as follows:
\begin{equation}
	\label{deqn_ex4}
	\hat{{\bf{F}}}=({\bf{F}}_{S}\times {\bf{F}}_{C}) \odot {\bf{F}} + {\bf{F}},
\end{equation}
where $\odot$ denotes pixel-by-pixel multiplication.

\subsubsection{Multi-level Conversion Module}
Feature fusion\cite{chaib2017deep,akilan2017late,yan2020blind} is an important way to make full use of the information from each individual feature input. Generally, the low-level features in shallow layers have higher resolution and contain more detailed information, while the high-level features in deep layers have lower resolution and stronger semantic representation ability. In addition, it has been demonstrated that the HVS tends to perform multi-level feature fusion in perceiving image quality \cite{wang2003multi}. In our method, we design a multi-level conversion module to fuse the low-level and high-level features from the $1^{st}$ layer and the $4^{th}$ layer of the backbone network to exploit the complementarity between them.

The multi-view projected images are obtained from the point cloud under different viewpoints. Therefore, there is a certain inter-dependency between each other. To capture and exploit such a kind of dependency, we build a graph based on the correlation between those multi-view projected images by taking each projected image feature representation as a node in the graph. As a consequence, the obtained multi-level attentive feature maps need to be converted into the feature vector ${\bf{h}}_{v_i} \in\mathbb{R}^{D}$ of the graph node $v_i$ ($i\in\{1,2,…,N\}$) by the multi-level conversion module, where $D$ is the feature dimension. Formally, the conversion process of a feature vector can be expressed as
\begin{equation}
	\label{deqn_ex5}
	{\bf{h}}_v = avg(\hat{{\bf{F}}}^{(1)}) \oplus avg(\hat{{\bf{F}}}^{(4)}),
\end{equation}
where $\hat{{\bf{F}}}^{(1)}$ and $\hat{{\bf{F}}^{(4)}}$ represent the attentive feature maps output by the attention block, and the input of the attention block comes from the $1^{st}$ and $4^{th}$ layers of the pre-trained backbone, respectively,  $\oplus$ represents the concatenation of the channels of the feature maps. Finally, we create a set of nodes ${\bf{V}}=[{\bf{h}}_{v_1},{\bf{h}}_{v_2},\cdots,{\bf{h}}_{v_N}] ^\mathrm{T}$ based on all the feature vectors, Meanwhile, the adjacency relation between any two nodes {$v_i$, $v_j$} can be expressed as
\begin{equation}
	\label{deqn_ex6}
	{{\bf{A}}(v_i,v_j)} = \begin{cases}
		1,&{\text{if }} RSDist(v_i,v_j) \le \theta \\ 
		{0,}&{\text{otherwise}} 
	\end{cases},
\end{equation}
\noindent where ${\bf{A}}\in\mathbb{R}^{N\times N}$ is the adjacency matrix representing the correlation between nodes of the graph, $RSDist(\cdot)$ computes the total RS of the projected images corresponding to the two nodes, and the RS threshold $\theta$ is set to $36^\circ$ with experiments. Specifically, if the RS is less than or equal to $\theta$, the two projected images are considered to be spatially connected and share some common information, otherwise they are not adjacent.
Then, by multiplying both sides of the adjacency matrix ${\bf{A}}$ by the square root of the degree matrix for normalization\cite{kipf2016semi,wu2019spatial}, the graph nodes with many neighbor nodes are avoided to have too much influence. The normalization formula is as follows
\begin{equation}
	\label{deqn_ex7}
	\widehat{{\bf{A}}} = {\bf{D}}^{-\frac{1}{2}} ({\bf{A}} + {\bf{I}}) {\bf{D}}^{\frac{1}{2}},
\end{equation}
\noindent where $\widehat{{\bf{A}}}$ is the normalized adjacency matrix. ${\bf{D}}$ is the degree matrix, which takes the degree of the corresponding node as the value only on the diagonal and is 0 in the rest of the positions. Concretely, ${\bf{D}}_{ii}=\sum_{j=0}^{N} {\bf{A}}(v_i,v_j)$ ($i\in\{1,2,\cdots,N\}$). ${\bf{I}}$ is the identity matrix and adds self-join to the adjacency matrix. Finally, we construct the graph ${\bf{G}}=({\bf{V}},\widehat{{\bf{A}}})$, so that the correlation between nodes $v_i$ and $v_j$ can be represented by the corresponding value ${{\bf{A}}(v_i,v_j)}$ in the adjacency matrix.

\subsection{GCN-based Quality Prediction}
After graph construction, we use GCN to model the interaction between the contents of different projected 2D images according to the graph ${\bf{G}}$, thus completing the quality prediction.
\subsubsection{Graph Convolutional Network}
We take the two graphs corresponding to the horizontal and vertical projection image feature groups into two GCNs without weight sharing, respectively, and update the new node representations by constantly exchanging neighborhood information based on the $\widehat{{\bf{A}}}$. The GCN consists of four graph convolutional blocks, and the number of output channels of these blocks are [512, 128, 32, 1]. The graph convolutional block include a graph convolutional layer, a softplus activation function, and a batch normalization layer. The process of GCN can be described as
\begin{equation}
	\label{deqn_ex8}
	({\bf{H}}^{(1)},{\bf{H}}^{(2)},{\bf{H}}^{(3)},{\bf{H}}^{(4)}) = M_G({\bf{G}},\widehat{{\bf{A}}};{\bf{w}}_G),
\end{equation}
\noindent where ${\bf{H}}^{(l)}$ is the feature matrix after activation of the $l^{th}$ layer of GCN, ${\bf{H}}^{(0)}={\bf{V}}$, $M_G$ denotes the GCN, ${\bf{w}}_G$ is the network parameter that is constantly updated during training. The layer-wise propagation rule of GCN is defined as follows
\begin{equation}
	\label{deqn_ex9}
	{\bf{H}}^{(l+1)} = \sigma(BN(\widehat{{\bf{A}}} {\bf{H}}^{(l)} {\bf{w}}_G^{(l)})),
\end{equation}
\noindent where $\sigma$ represents the softplus activation function, $BN$ represents batch normalization, ${\bf{w}}_G^{(l)}$ is the trainable weight matrix of the $l^{th}$ layer, and the size of the matrix is related to the number of input and output channels. Based on the above propagation rules, the GCN continuously learns the dependencies between nodes and uses the information in the adjacency matrix to aggregate the features of itself and its neighbors to extract richer features.

\subsubsection{Quality Prediction}
Here, we fuse the multi-level feature matrices ${\bf{H}}^{(l)} (l\in\{1,2,3,4\})$ output by GCN. Above all, the first dimension of the first three feature matrices is averaged pooling, so as to imitate the HVS to aggregate the feature information of different projection images. Then, the 1-dimensional perceptual feature matrix is obtained through the dimension reduction of the fully connected layer. Since the number of channels of the feature matrix output by the last layer is already 1, the average pooling layer and the fully connected layer are used for the feature information of the first dimension of the matrix, so as to enhance the diversity of features. The final obtained multi-level fusion feature matrix $\bar{H}$ is described as follows
\begin{equation}
\begin{aligned}
	\label{deqn_ex10}
	\bar{{\bf{H}}} = & L(\alpha({\bf{H}}^{(1)}))\oplus L(\alpha({\bf{H}}^{(2)})) \oplus 
                \\ & L(\alpha({\bf{H}}^{(3)})) \oplus L({\bf{H}}^{(4)})\oplus\alpha({\bf{H}}^{(4)}),
\end{aligned}
\end{equation}
\noindent in which $\alpha(\cdot)$ represents average pooling, $L(\cdot)$ represents the fully connected layer. From this, we extract the multi-level fusion feature matrix of horizontal and vertical projection image groups respectively, which are denoted as $\bar{{\bf{H}}}_H$ and $\bar{{\bf{H}}}_V$. The two groups of features from different projection directions have different detail information and can complement each other. Finally, after fusing the two feature matrices, we use the fully connected layer to automatically assign weights to $\bar{{\bf{H}}}_H$ and $\bar{{\bf{H}}}_V$ to predict the perceptual quality score of the point cloud.

\subsection{Network Training}
For the whole network, we simultaneously input 20 images from the two projected image groups to jointly optimize the two branches. The loss function used to optimize the model is $l_1$, which can be defined as
\begin{equation}
    \label{deqn_ex11}
    l_1=\frac{1}{n} \sum_{i=0}^{n} \lvert y_{i}-\bar{y}_{i}\rvert,
\end{equation}
\begin{equation}
    \label{deqn_ex12}
    \bar{y}_{i}=Q({\bf{P}},\widehat{{\bf{A}}};{\bf{w}}_F),
\end{equation}
\noindent where $n$ indicates batch size, $y_{i}$ and $\bar{y}_{i}$ indicates the $i^{th}$ subjective quality score and objective prediction score in the batch respectively. $\bar{y}_{i}$ is extracted through the GC-PCQA network $Q$. ${\bf{w}}_F$ is the trainable parameters of the network, which are updated by minimizing $l_1$.

\begin{table*}
\centering
\renewcommand\arraystretch{1.2}
\caption{Performance comparison results on SJTU and WPC databases. All indicators adopt absolute values for performance comparison for better visibility. The first, second, and third of the four indicators are marked in \textcolor{red}{red}, \textcolor{blue}{blue} and \textcolor{green}{green}, respectively.}  
\label{table1}  % 用于索引表格的标签
\resizebox{\linewidth}{!}{
\begin{tabular}{c|c|c|cccc|cccc}
\hline
\multirow{2}{*}{Ref} & \multirow{2}{*}{Type} & \multirow{2}{*}{Metric} & \multicolumn{4}{c|}{SJTU}                           & \multicolumn{4}{c}{WPC}   \\ \cline{4-11} 
& & & SRCC$\uparrow$ & PLCC$\uparrow$ & KRCC$\uparrow$ & RMSE$\downarrow$ & SRCC$\uparrow$ & PLCC$\uparrow$             & KRCC$\uparrow$ & RMSE$\downarrow$ \\ \hline
\multirow{15}{*}{FR} & \multirow{11}{*}{PC-Based}        
                     & $PSNR_{MSE,p2po}$        & 0.6002                    & 0.7622                    & 0.4917                    & 1.4382                    & 0.1607                    & 0.2673                    & 0.1147                    & 20.6947                    \\
                     &                                   
                     & $PSNR_{MSE,p2pl}$         & 0.5505                    & 0.7381                    & 0.4375                    & 1.5357                    & 0.1182                    & 0.2879                    & 0.0851                    & 21.1898                    \\
                     &                                   
                     & $PSNR_{HF,p2po}$         & 0.6744                    & 0.7737                    & 0.5217                    & 1.4481                    & 0.0557                    & 0.3555                    & 0.0384                    & 20.8197                    \\
                     &                                   
                     & $PSNR_{HF,p2pl}$         & 0.6208                    & 0.7286                    & 0.4701                    & 1.6000                    & 0.0989                    & 0.3263                    & 0.0681                    & 21.11                      \\
                     &                                   
                     & $AS_{Mean}$               & 0.5317                    & 0.5297                    & 0.3723                    & 2.7129                    & 0.2484                    & 0.3397                    & 0.1801                    & 21.5013                    \\
                     &                                   
                     & $AS_{RMS}$                & 0.5653                    & 0.7156                    & 0.4144                    & 1.6550                    & 0.2479                    & 0.3347                    & 0.1802                    & 21.5325                    \\
                     &                                   
                     & $AS_{MSE}$               & 0.5472                    & 0.5115                    & 0.3865                    & 2.6431                    & 0.2484                    & 0.3397                    & 0.1801                    & 21.5013                    \\
                     &                                   
                     & $PSNR_Y$                 & 0.7871                    & 0.8124                    & 0.6116                    & 1.3222                    & 0.5823                    & 0.6166                    & 0.4164                    & 17.9001                    \\
                     &                                   
                     & PCQM                    & 0.7748                    & 0.8301                    & 0.6152                    & 1.2978                    & 0.5504                    & 0.6162                    & 0.4409                    & 17.9027                    \\
                     &                                   
                     & PointSSIM               & 0.7051                    & 0.7422                    & 0.5321                    & 1.5601                    & 0.4639                    & 0.5225                    & 0.3394                    & 19.3863                    \\
                     &                                   
                     & GraphSIM                & \textcolor{blue}{0.8853}  & \textcolor{blue}{0.9158}  & \textcolor{green}{0.7063} & \textcolor{blue}{0.9462}  & 0.6217                    & 0.6833                    & 0.4562                    & 16.5107                    \\ \cline{2-11} 
                     & \multirow{4}{*}{Projection-Based} 
                     & SSIM                    & 0.8667                    & 0.8868                    & 0.6988                    & 1.0454                    & 0.6483                    & 0.6690                    & 0.4685                    & 16.8841                    \\
                     &                                   
                     & MS-SSIM                 & \textcolor{green}{0.8738} & 0.8930                    & \textcolor{blue}{0.7069}  & \textcolor{green}{1.0091} & 0.7179                    & 0.7349                    & 0.5385                    & 15.3341                    \\
                     &                                   
                     & IW-SSIM                 & 0.8638                    & 0.8932                    & 0.6934                    & 1.0268                    & \textcolor{blue}{0.7608}  & \textcolor{blue}{0.7688}  & \textcolor{blue}{0.5707}  & \textcolor{blue}{14.5453}  \\
                     &                                   
                     & VIFP                    & 0.8624                    & \textcolor{green}{0.8977} & 0.6934                    & 1.0173                    & \textcolor{green}{0.7426} & \textcolor{green}{0.7508} & \textcolor{green}{0.5575} & \textcolor{green}{15.0328} \\ \hline
RR                   & PC-Based                          
                     & PCMRR                   & 0.5622                    & 0.6699                    & 0.4091                    & 1.7589                    & 0.3605                    & 0.3926                    & 0.2543                    & 20.9203                    \\ \hline
\multirow{5}{*}{NR}  & \multirow{2}{*}{PC-Based}         
                     & 3D-NSS                  & 0.7819                    & 0.7813                    & 0.6023                    & 1.7740                    & 0.6309                    & 0.6284                    & 0.4573                    & 18.1706                    \\
                     &                                   
                     & ResSCNN                 & 0.8328                    & 0.8865                    & 0.6514                    & 1.0728                    & 0.4362                    & 0.4531                    & 0.2987                    & 20.2591                    \\ \cline{2-11} 
                     & \multirow{3}{*}{Projection-Based} 
                     & PQANet                  & 0.7593                    & 0.7998                    & 0.5796                    & 1.3773                    & 0.6368                    & 0.6671                    & 0.4684                    & 16.6758                    \\
                     &                                   
                     & IT-PCQA                 & 0.8286                    & 0.8605                    & 0.6453                    & 1.1686                    & 0.4329                    & 0.4870                    & 0.3006                    & 19.896                     \\
                     &                                   
                     & \bf{GC-PCQA (Ours)}               & \textcolor{red}{0.9108}   & \textcolor{red}{0.9301}   & \textcolor{red}{0.7546}   & \textcolor{red}{0.8691}   & \textcolor{red}{0.8054}   & \textcolor{red}{0.8091}   & \textcolor{red}{0.6246}   & \textcolor{red}{13.3405}   \\ \hline
\end{tabular}}
\end{table*}

\section{Experimental Results and Analysis}
In this section, we first introduce the adopted subject-rated databases
and performance measures. Then, the implementation details are provided. Finally, we conduct extensive experiments and analyze the results to verify our proposed method, including both performance comparison and ablation study.

\subsection{Databases and Performance Measures}
\subsubsection{Databases}
We perform experiments on two publicly available 3D point cloud databases which consist of SJTU\cite{yang2020predict} and WPC\cite{su2019perceptual}.

The SJTU database includes nine pristine and 378 distorted point clouds generated from seven distortion types. Each distortion type corresponds to six distortion levels. 
The subjective testing protocol uses Absolute Category Rating (ACR), and the subjective scores are in the form of MOS values ranging from 1 to 10.

The WPC database has 20 original point clouds. For each reference point cloud, 37 distorted point clouds are created by simulating five distortion types (i.e., Downsample, Gaussian white noise, G-PCC(T), V-PCC, G-PCC(O)), leading to 740 distorted point clouds in total. Each distorted point cloud also relates to a MOS value. 
The subjective testing protocol uses the Double Stimulus Impairment Scale (DSIS), with MOS values ranging from 0 to 100.

\subsubsection{Performance Measures}
We apply four measures to evaluate and compare different PCQA methods, including Spearman’s Rank Correlation Coefficient (SRCC), Pearson’s linear correlation coefficient (PLCC), Kendall Rank Correlation Coefficient (KRCC), and root mean squared error (RMSE). The SRCC and KRCC are used to measure the monotonicity, while PLCC and RMSE are used to evaluate the accuracy. Higher correlation coefficients and lower RMSE represent better performance. 
To resolve the scale differences between the predicted quality scores and the subjective scores, we use a five-parameter logistic function \cite{video2003final} after calculating the initial prediction results, which can be expressed as
\begin{equation}
    \label{deqn_ex14}
    y=\beta_1(\frac{1}{2}-\frac{1}{1+\text{exp}(\beta_2(x-\beta_3))})+\beta_4x+\beta_5,
\end{equation}
\noindent where $x$ is the initial prediction result of the PCQA metric, $y$ indicates the mapped objective quality score through the five-parameter logistic function, and $\beta_i$ $(i\in\{1,2,\cdots,5\})$ are the fitting parameters.

\subsection{Implementation Details}
In our experiments, we employ PyTorch as the deep learning framework and the computer operating system is Ubuntu18.04. Moreover, the GPU is used to accelerate the training and testing procedures. The adaptive moment estimation optimizer (Adam)\cite{kingma2014adam} is used for model training. We set batch size and initial learning rate as 32 and 1e-3, respectively. Additionally, the learning rate is reduced to 0.5 times of the original one every 10 epochs until the final convergence.

The network $F$ is trained for 50 epochs and the training terminates early when there is no further optimized $w_F$ for 20 epochs. Meanwhile, we exploit data augmentation methods such as horizontal and vertical flipping to enhance the generalization ability of the proposed network.

In addition, the k-fold cross-validation strategy is used for the performance test. For each point cloud quality database, $\frac{K-1}{K}$ distorted samples are randomly selected from the database as the train sets, and the rest point clouds are used as the test sets. Specifically, we choose $K$ equalling to 9 and 5 for the SJTU and WPC databases, respectively. The final results can be obtained by averaging the performance values from $K$ times.

\begin{figure*}[t]
	\centering
	\includegraphics[width=\linewidth]{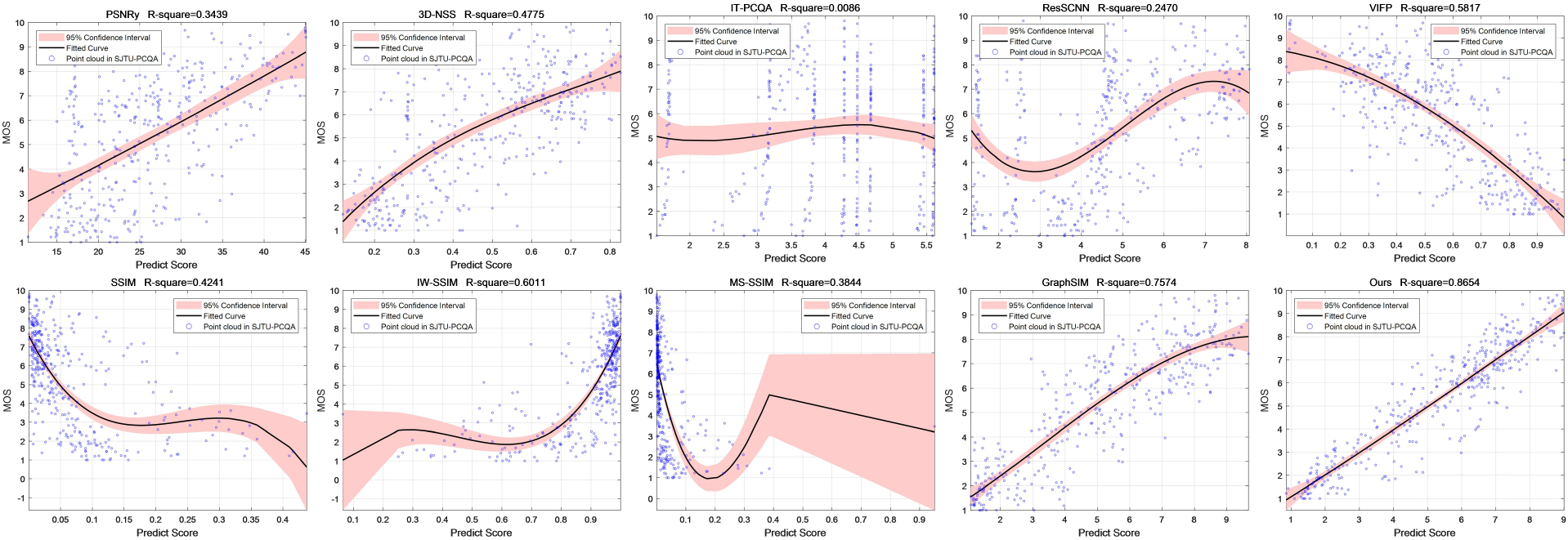}
    \includegraphics[width=\linewidth]{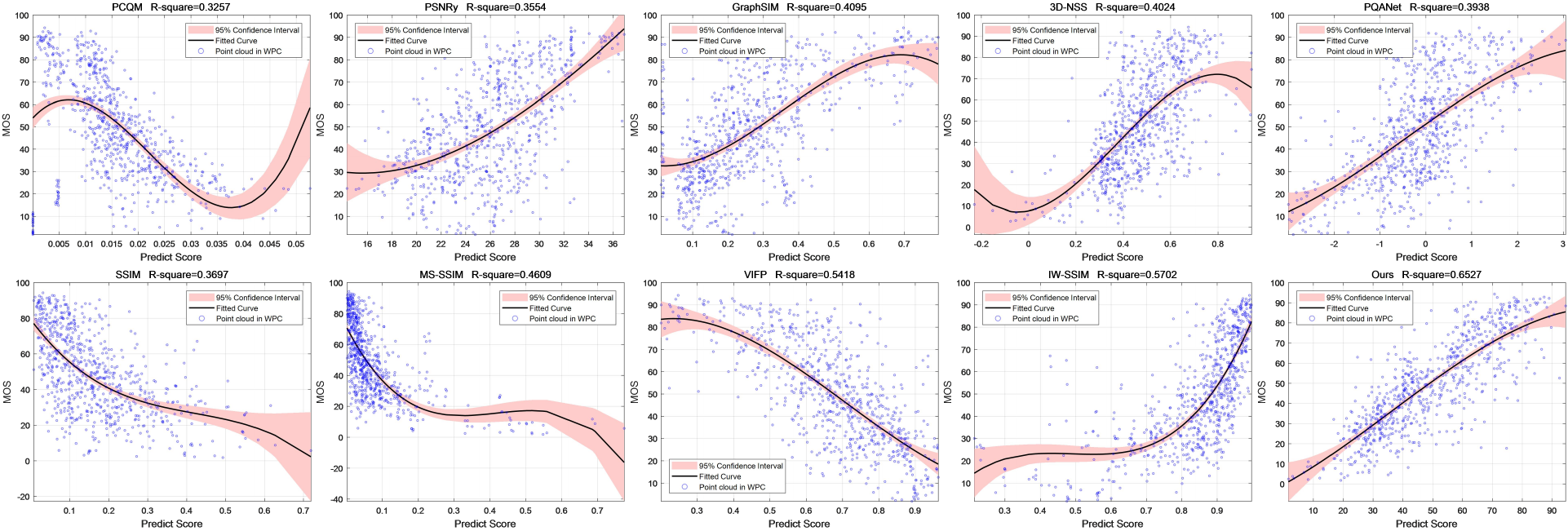}
	\caption{Scatter plot between objective prediction scores and MOS for the top ten PCQA metrics in the experiment. The X-axis is the objective prediction score of the PCQA metric, and the Y-axis is the corresponding MOS. The first ten figures are the results on SJTU database, from top left to bottom right are $PSNR_Y$\cite{mekuria2017per}, 3D-NSS\cite{zhang2022no}, IT-PCQA\cite{yang2022no}, ResSCNN\cite{liu2023point}, VIFP\cite{sheikh2006image}, SSIM\cite{wang2004image}, IW-SSIM\cite{wang2010info}, MS-SSIM\cite{wang2003multi}, GraphSIM\cite{diniz2020local} and our proposed method. The last ten figures are the results on the WPC database, from top left to bottom right are PCQM\cite{meynet2020pcqm}, $PSNR_Y$\cite{mekuria2017per}, GraphSIM\cite{diniz2020local}, 3D-NSS\cite{zhang2022no}, PQANet\cite{liu2021pqa}, SSIM\cite{wang2004image}, MS-SSIM\cite{wang2003multi}, VIFP\cite{sheikh2006image}, IW-SSIM\cite{wang2010info} and our proposed method. R-Square score, 95\% confidence interval, and fitted curve are calculated for each scatter plot.}
	\label{fig_4}
\end{figure*}

\subsection{Performance Comparison}
We compare our proposed GC-PCQA with 20 state-of-the-art quality assessment methods. As mentioned in Section II, existing PCQA metrics can be divided into two types: PC-based metrics and projection-based metrics. PC-based metrics are directly evaluated from 3D point clouds, including $PSNR_{MSE,p2po}$\cite{mekuria2016eval}, $PSNR_{HF,p2po}$\cite{mekuria2016eval}, $PSNR_{MSE,p2pl}$\cite{tian2017geometric}, $PSNR_{HF,p2pl}$\cite{tian2017geometric}, $AS_{Mean}$\cite{alexiou2018point}, $AS_{RMS}$\cite{alexiou2018point}, $AS_{MSE}$\cite{alexiou2018point}, $PSNR_Y$\cite{mekuria2017per}, PCQM\cite{meynet2020pcqm}, PointSSIM\cite{alexiou2020to}, GraphSIM\cite{diniz2020local}, PCMRR\cite{viola2020red}, 3D-NSS\cite{zhang2022no}, and ResSCNN\cite{liu2023point}. The projection-based metrics operate on the projected 2D images of point clouds, including SSIM\cite{wang2004image}, MS-SSIM\cite{wang2003multi}, IW-SSIM\cite{wang2010info}, VIFP\cite{sheikh2006image}, PQANet\cite{liu2021pqa}, and IT-PCQA\cite{yang2022no}. It is worth mentioning that the performance results of SSIM, MS-SSIM, IW-SSIM, and VIFP are the average values from six perpendicular projections \cite{yang2020predict,zhou2023reduced}.

\begin{table*}[t!]
\centering
\renewcommand\arraystretch{1.5}
\caption{SRCC performance evaluation of existing PCQA metrics based on point cloud content and distortion type is performed on the SJTU database. Absolute SRCC is used for comparison to obtain better visibility. The letters A-R in the table stand for $PSNR_{MSE,p2po}$, $PSNR_{MSE,p2pl}$, $PSNR_{HF,p2po}$, $PSNR_{HF,p2pl}$, $AS_{RMS}$, $PSNR_Y$, PCQM, PointSSIM, GraphSIM, SSIM, MS-SSIM, IW-SSIM, VIFP, PCMRR, 3D-NSS, ResSCNN, PQANet, IT-PCQA and our proposed method in turn. The first, second, and third places in the SPCC indicator are marked in \textcolor{red}{red}, \textcolor{blue}{blue} and \textcolor{green}{green}, respectively.} 
\label{table2}  
\resizebox{\linewidth}{!}{
\begin{tabular}{cc|ccccccccccccc|c|ccccc}
\hline
\multicolumn{2}{c|}{\multirow{2}{*}{Subset}} &\multicolumn{13}{c|}{FR} &RR &\multicolumn{5}{c}{NR}  \\ \cline{3-21} 
\multicolumn{2}{c|}{} &A &B &C &D &E &F &G &H &I &J &K &L &M &N &O &P &Q &R &GC-PCQA  \\ \hline
\multicolumn{1}{c|}{\multirow{9}{*}{\rotatebox{90}{Content}}}    
& Redandblack  & 0.6196 & 0.5943 & 0.7421 & 0.6819 & 0.5799 & 0.7478 & 0.8024 & 0.6670 & 0.8702 & 0.8603 & 0.8718 & \textcolor{blue}{0.8911} & \textcolor{green}{0.8885} & 0.6506 & 0.8647 & 0.8003 & 0.8603 & 0.8557 & \textcolor{red}{0.9057} \\
\multicolumn{1}{c|}{}                            
& Romanoillamp & 0.4247 & 0.3617 & 0.7457 & 0.6032 & 0.6022 & 0.4278 & 0.5145 & 0.5150 & \textcolor{blue}{0.8525} & 0.7509 & 0.7869 & \textcolor{green}{0.7939} & 0.7882 & 0.6044 & 0.6885 & 0.6193 & 0.7509 & 0.7248 & \textcolor{red}{0.9041} \\
\multicolumn{1}{c|}{}                            
& Loot         & 0.6738 & 0.6405 & 0.7447 & 0.6391 & 0.4817 & 0.7875 & 0.8426 & 0.7299 & \textcolor{green}{0.8868} & 0.8693 & 0.8809 & 0.8846 & 0.8619 & 0.6770 & \textcolor{blue}{0.8890} & 0.8780 & 0.8693 & 0.8778 & \textcolor{red}{0.9481} \\
\multicolumn{1}{c|}{}                            
& Soldier      & 0.6781 & 0.6478 & 0.7493 & 0.6329 & 0.5404 & 0.8336 & 0.8684 & 0.7718 & \textcolor{green}{0.9118} & 0.8917 & 0.8843 & 0.8843 & 0.8744 & 0.5809 & 0.8731 & \textcolor{blue}{0.9123} & 0.8917 & 0.8050 & \textcolor{red}{0.9253} \\
\multicolumn{1}{c|}{}                            
& ULB Unicorn  & 0.7085 & 0.6082 & 0.8500 & 0.8081 & 0.4773 & 0.8687 & 0.7496 & 0.5715 & 0.8597 & \textcolor{green}{0.9084} & 0.8981 & 0.8548 & 0.8514 & 0.5148 & 0.4101 & 0.8364 & \textcolor{green}{0.9084} & \textcolor{red}{0.9129} & \textcolor{blue}{0.9109} \\
\multicolumn{1}{c|}{}                            
& Longdress    & 0.6640 & 0.6437 & 0.7885 & 0.7096 & 0.5704 & \textcolor{green}{0.9326} & 0.8896 & 0.8608 & \textcolor{red}{0.9499} & 0.9245 & 0.9191 & 0.8710 & 0.8976 & 0.6474 & 0.9005 & 0.8650 & 0.9245 & 0.8243 & \textcolor{blue}{0.9441} \\
\multicolumn{1}{c|}{}                            
& Statue       & 0.5678 & 0.5362 & 0.5883 & 0.5652 & 0.6291 & 0.8241 & 0.7483 & 0.7391 & \textcolor{green}{0.8744} & 0.8578 & 0.8663 & 0.8428 & 0.8637 & 0.4181 & 0.8520 & \textcolor{red}{0.9002} & 0.8578 & \textcolor{blue}{0.8757} & 0.8633 \\
\multicolumn{1}{c|}{}                            
& Shiva        & 0.4129 & 0.4074 & 0.1168 & 0.2689 & 0.7057 & 0.8375 & 0.8060 & 0.7896 & 0.8595 & \textcolor{red}{0.8968} & \textcolor{blue}{0.8914} & 0.8744 & \textcolor{green}{0.8903} & 0.4884 & 0.8198 & 0.8599 & \textcolor{red}{0.8968} & 0.8243 & 0.8866 \\
\multicolumn{1}{c|}{}                            
& Hhi          & 0.6526 & 0.5150 & 0.7443 & 0.6785 & 0.5012 & 0.8242 & 0.7524 & 0.7010 & \textcolor{blue}{0.9028} & 0.8409 & 0.8658 & \textcolor{green}{0.8773} & 0.8462 & 0.4785 & 0.7394 & 0.8240 & 0.8409 & 0.7577 & \textcolor{red}{0.9089} \\ \hline
\multicolumn{1}{c|}{\multirow{7}{*}{\rotatebox{90}{Distortion}}} 
& OT           & 0.4407 & 0.4407 & 0.3788 & 0.3524 & 0.5210 & 0.3068 & 0.6495 & \textcolor{blue}{0.7108} & \textcolor{green}{0.7049} & 0.2198 & 0.2712 & 0.3382 & 0.3743 & 0.1800 & 0.4068 & 0.1683 & 0.0883 & 0.0189 & \textcolor{red}{0.8892} \\
\multicolumn{1}{c|}{}                            
& CN           & NaN    & NaN    & NaN    & NaN    & NaN    & 0.5588 & 0.6070 & \textcolor{green}{0.7660} & \textcolor{blue}{0.7779} & 0.6283 & 0.6453 & 0.7531 & 0.7429 & 0.7157 & 0.1480 & 0.2265 & 0.5507 & 0.0655 & \textcolor{red}{0.9021} \\
\multicolumn{1}{c|}{}                            
& DS           & 0.4495 & 0.4489 & 0.6847 & 0.3286 & 0.3653 & 0.4697 & 0.6990 & \textcolor{green}{0.8500} & \textcolor{blue}{0.8654} & 0.3246 & 0.4718 & 0.4535 & 0.4546 & 0.1489 & 0.5051 & 0.4292 & 0.2958 & 0.0556 & \textcolor{red}{0.8918} \\
\multicolumn{1}{c|}{}                            
& D+C          & 0.5735 & 0.5979 & 0.7619 & 0.7499 & 0.4025 & 0.7397 & \textcolor{green}{0.8014} & 0.7449 & \textcolor{blue}{0.8846} & 0.5062 & 0.6281 & 0.6661 & 0.6932 & 0.6120 & 0.5895 & 0.5158 & 0.4899 & 0.0468 & \textcolor{red}{0.9500} \\
\multicolumn{1}{c|}{}                            
& D+G          & 0.6779 & 0.7058 & 0.7423 & 0.7196 & \textcolor{green}{0.8915} & 0.5413 & 0.7476 & \textcolor{blue}{0.9288} & 0.8833 & 0.6920 & 0.7589 & 0.8222 & 0.7989 & 0.7439 & 0.7442 & 0.5263 & 0.5033 & 0.0411 & \textcolor{red}{0.9664} \\
\multicolumn{1}{c|}{}                            
& GGN          & 0.7008 & 0.7144 & 0.7453 & 0.7328 & \textcolor{blue}{0.9376} & 0.5727 & 0.7143 & 0.9027 & \textcolor{green}{0.9064} & 0.7436 & 0.7783 & 0.8324 & 0.8436 & 0.7813 & 0.8435 & 0.4497 & 0.3771 & 0.0798 & \textcolor{red}{0.9546} \\
\multicolumn{1}{c|}{}                            
& C+G          & 0.7577 & 0.7758 & 0.8205 & 0.8025 & \textcolor{green}{0.9241} & 0.6692 & 0.7078 & 0.7991 & \textcolor{blue}{0.9334} & 0.7307 & 0.7948 & 0.8406 & 0.8463 & 0.8329 & 0.8645 & 0.5523 & 0.6137 & 0.1044 & \textcolor{red}{0.9681} \\ \hline
\end{tabular}}
\end{table*}

\begin{table*}[!t]
\centering
\renewcommand\arraystretch{1.5}
\caption{SRCC performance evaluation of existing PCQA metrics based on point cloud content and distortion type is performed on the WPC database. Absolute SRCC is used for comparison to obtain better visibility. The letters represent the same PCQA metrics as the table above. The first, second, and third places in the SPCC indicator are marked in \textcolor{red}{red}, \textcolor{blue}{blue} and \textcolor{green}{green}, respectively.} 
\label{table3}  % 用于索引表格的标签
\resizebox{\linewidth}{!}{
\begin{tabular}{cc|ccccccccccccc|c|ccccc}
\hline
\multicolumn{2}{c|}{\multirow{2}{*}{Subset}} &\multicolumn{13}{c|}{FR} &RR &\multicolumn{5}{c}{NR} \\ \cline{3-21} 
\multicolumn{2}{c|}{} &A &B &C &D &E &F &G &H &I &J &K &L &M &N &O &P &Q &R &GC-PCQA \\ \hline
\multicolumn{1}{c|}{\multirow{20}{*}{\rotatebox{90}{Content}}}   
& Bag            & 0.6669 & 0.5751 & 0.4363 & 0.4365 & 0.4325 & \textcolor{red}{0.8051} & 0.5955 & 0.4829 & 0.7164 & 0.7300 & 0.7584 & 0.7309 & 0.7093 & 0.6069 & \textcolor{blue}{0.7731} & 0.1603 & 0.3504 & 0.6174 & \textcolor{green}{0.7587} \\
\multicolumn{1}{c|}{}                            
& Banana         & 0.6471 & 0.5691 & 0.1933 & 0.2033 & 0.3147 & 0.6211 & 0.4649 & 0.2202 & 0.5045 & \textcolor{red}{0.8011} & 0.7677 & \textcolor{blue}{0.7790} & \textcolor{green}{0.7771} & 0.5287 & 0.6524 & 0.2475 & 0.6949 & 0.2485 & 0.5503 \\
\multicolumn{1}{c|}{}                            
& Biscuits       & 0.5252 & 0.4160 & 0.3085 & 0.3368 & 0.3505 & 0.7764 & 0.6245 & 0.5816 & 0.7198 & \textcolor{blue}{0.9173} & \textcolor{red}{0.9500} & \textcolor{green}{0.7992} & 0.7416 & 0.4310 & 0.6645 & 0.4765 & 0.6147 & 0.3570 & 0.7468 \\
\multicolumn{1}{c|}{}                            
& Cake           & 0.3074 & 0.1798 & 0.1724 & 0.1796 & 0.0609 & 0.5180 & 0.4566 & 0.3177 & 0.4251 & \textcolor{green}{0.7390} & \textcolor{blue}{0.7691} & 0.6534 & 0.6477 & 0.3070 & 0.4547 & 0.4467 & 0.5835 & 0.7300 & \textcolor{red}{0.8532} \\
\multicolumn{1}{c|}{}                            
& Cauliflower    & 0.3501 & 0.2058 & 0.0918 & 0.1653 & 0.1781 & 0.5927 & 0.4903 & 0.4237 & 0.5529 & 0.8004 & \textcolor{blue}{0.8608} & \textcolor{green}{0.8182} & 0.7008 & 0.4187 & 0.5517 & 0.5095 & 0.6238 & 0.0593 & \textcolor{red}{0.9239} \\
\multicolumn{1}{c|}{}                            
& Flowerpot      & 0.6509 & 0.5298 & 0.4348 & 0.4515 & 0.3629 & 0.6385 & 0.5875 & 0.3784 & 0.6609 & 0.8303 & \textcolor{red}{0.9066} & \textcolor{blue}{0.9047} & \textcolor{green}{0.8954} & 0.0477 & 0.6958 & 0.4900 & 0.2357 & 0.8127 & 0.7591 \\
\multicolumn{1}{c|}{}                            
& GlassesCase    & 0.5845 & 0.4390 & 0.2020 & 0.3238 & 0.4288 & \textcolor{blue}{0.7826} & 0.5861 & 0.5258 & 0.6546 & 0.7617 & 0.7577 & 0.7304 & 0.7459 & 0.3883 & 0.4790 & 0.2003 & 0.7674 & \textcolor{green}{0.7750} & \textcolor{red}{0.8826} \\
\multicolumn{1}{c|}{}                            
& HoneydewMelon  & 0.4890 & 0.3299 & 0.2768 & 0.2300 & 0.3228 & 0.6740 & 0.4500 & 0.5609 & 0.7248 & \textcolor{green}{0.8549} & \textcolor{blue}{0.8917} & \textcolor{red}{0.9180} & 0.8279 & 0.5742 & 0.7229 & 0.4026 & 0.7418 & 0.7352 & 0.7387 \\
\multicolumn{1}{c|}{}                            
& House          & 0.5866 & 0.4483 & 0.3429 & 0.3434 & 0.4522 & \textcolor{green}{0.7798} & 0.5880 & 0.5590 & 0.7373 & 0.7788 & 0.7793 & 0.7357 & 0.7200 & 0.4905 & 0.7646 & 0.4780 & \textcolor{blue}{0.8668} & 0.4201 & \textcolor{red}{0.9343} \\
\multicolumn{1}{c|}{}                            
& Litchi         & 0.5109 & 0.4291 & 0.3478 & 0.3204 & 0.3554 & 0.7027 & 0.5965 & 0.6422 & 0.6958 & 0.7748 & \textcolor{blue}{0.8623} & 0.7496 & 0.7018 & 0.4839 & \textcolor{green}{0.8113} & 0.1994 & 0.7207 & 0.0868 & \textcolor{red}{0.8879} \\
\multicolumn{1}{c|}{}                            & Mushroom       & 0.6396 & 0.5156 & 0.3486 & 0.3105 & 0.2911 & 0.6550 & 0.5725 & 0.5443 & 0.6802 & 0.7821 & \textcolor{red}{0.8781} & \textcolor{green}{0.8160} & 0.7897 & 0.2556 & 0.8153 & 0.0754 & 0.5835 & 0.3570 & \textcolor{blue}{0.8608} \\
\multicolumn{1}{c|}{}                            
& PenContainer   & 0.7720 & 0.6688 & 0.2159 & 0.3635 & 0.5465 & 0.7328 & 0.6394 & 0.5948 & 0.8250 & \textcolor{red}{0.8954} & \textcolor{green}{0.8758} & 0.8485 & 0.8397 & 0.6830 & 0.7809 & 0.5676 & 0.6470 & 0.7859 & \textcolor{blue}{0.8928} \\
\multicolumn{1}{c|}{}                            
& Pineapple      & 0.3777 & 0.2785 & 0.1376 & 0.1831 & 0.2155 & 0.7217 & 0.6427 & 0.5386 & 0.6401 & \textcolor{green}{0.7307} & \textcolor{blue}{0.7805} & 0.5856 & 0.6441 & 0.4011 & 0.6074 & 0.5275 & 0.6318 & 0.5913 & \textcolor{red}{0.8862} \\
\multicolumn{1}{c|}{}                            
& PingpongBat    & 0.5924 & 0.4984 & 0.4958 & 0.4357 & 0.4521 & 0.5428 & 0.5783 & 0.6051 & 0.7697 & \textcolor{green}{0.8054} & \textcolor{red}{0.8812} & 0.7570 & 0.7539 & 0.5092 & 0.6935 & 0.3518 & 0.6358 & 0.4737 & \textcolor{blue}{0.8760} \\
\multicolumn{1}{c|}{}                            
& PuerTea        & 0.6069 & 0.4746 & 0.1173 & 0.0384 & 0.4734 & 0.7639 & 0.5685 & 0.4139 & 0.7999 & \textcolor{red}{0.8917} & \textcolor{blue}{0.8668} & \textcolor{green}{0.8359} & 0.7866 & 0.4308 & 0.4763 & 0.1456 & 0.7359 & 0.5467 & 0.6584 \\
\multicolumn{1}{c|}{}                            
& Pumpkin        & 0.4947 & 0.3423 & 0.3092 & 0.3068 & 0.3220 & 0.6901 & 0.5934 & 0.5699 & 0.6517 & \textcolor{red}{0.9111} & 0.8156 & \textcolor{blue}{0.9042} & \textcolor{green}{0.8976} & 0.3241 & 0.5768 & 0.4052 & 0.7857 & 0.5536 & 0.8435 \\
\multicolumn{1}{c|}{}                            
& Ship           & 0.7464 & 0.6267 & 0.3404 & 0.5158 & 0.4943 & 0.7786 & 0.5434 & 0.4488 & 0.7558 & \textcolor{red}{0.8973} & \textcolor{blue}{0.8578} & \textcolor{green}{0.8340} & 0.8013 & 0.4400 & 0.6935 & 0.6612 & 0.5349 & 0.3777 & 0.8054 \\
\multicolumn{1}{c|}{}                            
& Statue         & 0.8040 & 0.6707 & 0.2450 & 0.4487 & 0.4900 & 0.7001 & 0.5714 & 0.5085 & 0.7390 & \textcolor{green}{0.8985} & \textcolor{red}{0.9372} & \textcolor{blue}{0.9099} & 0.8950 & 0.1811 & 0.6368 & 0.5782 & 0.3762 & 0.4976 & 0.8743 \\
\multicolumn{1}{c|}{}                            
& Stone          & 0.6219 & 0.5129 & 0.3551 & 0.3424 & 0.3649 & 0.7115 & 0.6475 & 0.6126 & 0.1920 & \textcolor{green}{0.8426} & \textcolor{red}{0.8881} & \textcolor{blue}{0.8587} & 0.8196 & 0.3632 & 0.6968 & 0.2122 & 0.8234 & 0.1790 & 0.8222 \\
\multicolumn{1}{c|}{}                            
& ToolBox        & 0.3937 & 0.2969 & 0.1972 & 0.1884 & 0.2984 & \textcolor{blue}{0.8706} & 0.6304 & 0.4927 & 0.7935 & 0.7821 & 0.8255 & 0.8056 & 0.7411 & 0.5239 & 0.5806 & 0.5026 & \textcolor{green}{0.8653} & 0.4694 & \textcolor{red}{0.9336} \\ \hline
\multicolumn{1}{c|}{\multirow{5}{*}{\rotatebox{90}{Distortion}}} 
& Downsampling   & 0.4815 & 0.3251 & 0.5356 & 0.4879 & 0.2465 & 0.5542 & 0.4537 & 0.8319 & 0.7903 & 0.8234 & \textcolor{red}{0.8834} & \textcolor{green}{0.8822} & \textcolor{blue}{0.8828} & 0.7407 & 0.7508 & 0.2899 & 0.7234 & 0.3327 & 0.8148 \\
\multicolumn{1}{c|}{}                            
& Gaussian noise & 0.6155 & 0.6194 & 0.6149 & 0.6150 & 0.6844 & 0.7644 & 0.8775 & 0.5844 & 0.7469 & 0.6264 & 0.7118 & \textcolor{blue}{0.8560} & \textcolor{red}{0.8847} & 0.7762 & 0.7460 & 0.5459 & 0.7938 & 0.1718 & \textcolor{green}{0.8533} \\
\multicolumn{1}{c|}{}                            
& G-PCC (T)      & 0.3451 & 0.3568 & 0.2811 & 0.3085 & 0.1342 & 0.5916 & \textcolor{blue}{0.7775} & 0.6745 & \textcolor{green}{0.7457} & 0.4669 & 0.6042 & 0.6742 & 0.6304 & 0.2702 & 0.5947 & 0.2531 & 0.4710 & 0.1987 & \textcolor{red}{0.8038} \\
\multicolumn{1}{c|}{}                            
& V-PCC          & 0.1602 & 0.1992 & 0.2051 & 0.2370 & 0.3877 & 0.3203 & 0.5534 & 0.3546 & 0.5989 & 0.5141 & 0.5812 & \textcolor{blue}{0.7063} & \textcolor{red}{0.7410} & 0.2966 & 0.3927 & 0.1028 & 0.0045 & 0.0090 & \textcolor{green}{0.6830} \\
\multicolumn{1}{c|}{}                            
& G-PCC (O)      & NaN    & NaN    & NaN    & NaN    & 0.0350 & 0.8072 & \textcolor{red}{0.8944} & 0.7917 & \textcolor{green}{0.8258} & 0.5290 & 0.7214 & 0.7128 & 0.7116 & 0.6468 & 0.2891 & 0.0247 & 0.4204 & 0.1180 & \textcolor{blue}{0.8874} \\ \hline
\end{tabular}}
\end{table*}

The performance comparison results on SJTU and WPC databases are shown in Table \ref{table1}. From the table, we can draw several conclusions: 
(1) Compared to existing quality assessment methods, the proposed GC-PCQA achieves the best performance on both databases, which demonstrates the effectiveness of the proposed method. To be specific, the SRCC score of GC-PCQA is 0.0255 higher than that of the second-place GraphSIM on the SJTU database, and 0.0446 higher than that of the second-place IW-SSIM on the WPC database. 
(2) Since the WPC database has more point cloud data and more complex distortions, the performance of PCQA metrics on the WPC database shows a significant degradation compared to that on the SJTU database. For example, ResSCNN performs well on the SJTU database, but its SRCC decreases by 0.397 on the WPC database. Compared with other comparison methods, our proposed metric achieves promising results on both databases without excessive performance degradation. Therefore, GC-PCQA has stronger learning ability and can maintain better performance in more complex distorted point cloud databases. 
(3) Although the WPC database is more challenging, the DSIS method allows subjects to directly perceive and compare quality differences, improving the objectivity and reliability of subjective scores. Ultimately, our method consistently achieves excellent results in complex scenarios, demonstrating its robust learning capabilities.
(4) Four and five of the top-five PCQA metrics belong to projection-based metrics for the SJTU and WPC databases, respectively. This proves the effectiveness of converting point clouds into multiple projected 2D images for quality assessment. That is, the projection can help evaluate the visual quality of 3D point clouds. 
To get a more intuitive understanding of the performance for different PCQA metrics, we draw a scatter plot regarding the predicted scores and MOS, as shown in Fig. \ref{fig_4}. In this figure, we adopt the fitted curve to describe the relationship between objective predictions and MOS values, and the closer to the diagonal line, the better. Besides, R-Square is used to evaluate the goodness of the model fit, and the value range is [0, 1]. The closer the value is to 1, the better the model fitted the data. It can be clearly observed that the proposed GC-PCQA achieves the best R-Square scores on both SJTU and WPC databases, indicating that the objective predictions of the model fit the MOS well.

\begin{figure}[t]
	\centering
	\includegraphics[width=\linewidth]{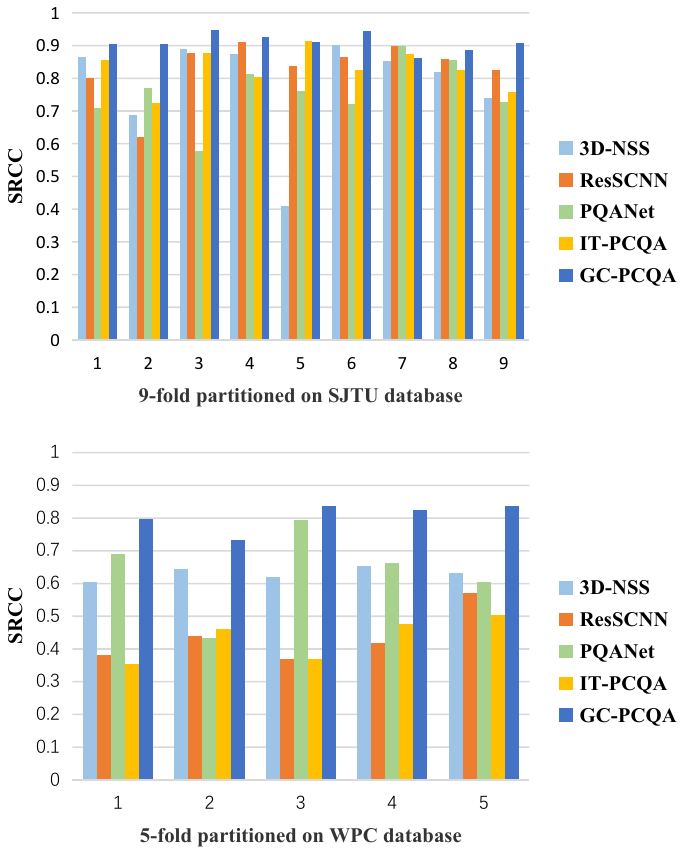}
	\caption{Performance comparison results of NR method in different partition test databases of SJTU and WPC. The SJTU database uses 9-fold cross-validation, while the WPC database uses 5-fold cross-validation.}
	\label{fig_5}
\end{figure}

Since point cloud quality databases involve various contents and distortion types, it is interesting to test the performance of existing PCQA metrics regarding each individual point cloud content and distortion type. We report the experimental results in Tables \ref{table2} and \ref{table3}. From the two tables, we can observe that: (1) On the smaller database subset, compared with the NR method, the FR method achieves more top-three performances due to the existence of the reference point cloud. (2) Our proposed method achieves the top-three SRCC performance for 14 times among the 16 subset experiments on the SJTU database, including the best performance 12 times. Meanwhile, in the 25 subset experiments on the WPC database, it achieves the top-three SRCC performance 15 times, including the best performance 8 times. Thus, our model can deliver superior performance, which further demonstrates the effectiveness of our proposed method. (3) The projection-based metric achieves the best performance 15 and 23 times for the SJTU and WPC databases. This also proves the advantages of the projection method in PCQA.

\begin{figure}[t]
	\centering
	\includegraphics[width=0.9\linewidth]{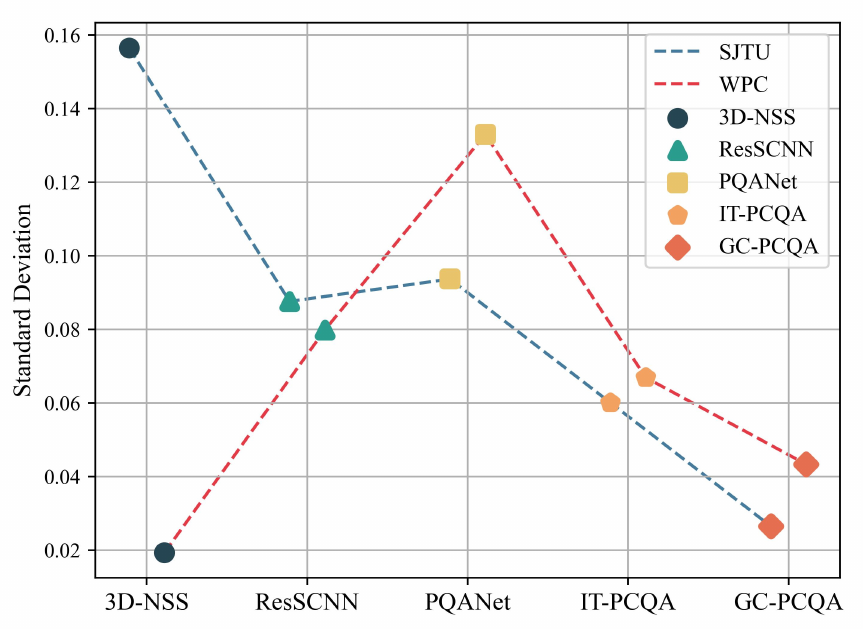}
	\caption{The results of the standard deviation of the mean SRCC performance for different PCQA metrics.}
	\label{fig_std}
\end{figure}

Considering that NR methods do not rely on original reference point clouds, they are more suitable in practical application scenarios. Therefore, here we perform k-fold cross-validation for NR methods and visualize all the results on the SJTU and WPC databases, as shown in Fig. \ref{fig_5}. The abscissa represents the test database with different partitions, and the ordinate represents the value of SRCC. From this figure, it can be observed that the SRCC results of our proposed method are better than the other NR methods on seven folds out of a total number of nine folds on the SJTU database, and the results of the remaining two folds are not far from the best performance. Moreover, it consistently outperforms other NR methods over all folds on the more complex WPC database. This shows that our proposed NR method can predict the perceptual quality of point clouds more accurately and maintain a certain performance in complex visual environments.
Additionally, we show the standard deviation of the mean SRCC of different PCQA metrics across all folds, as shown in Fig. \ref{fig_std}. The results indicate that our method achieves a lower standard deviation on both databases, showcasing a more stable performance.

\begin{table}[t]
\centering
\renewcommand\arraystretch{1.2}
\caption{Performance contribution results of projected images and GCN on SJTU and WPC databases. The best performance is indicated in \bf{BOLD}. }  % 表格标题
\label{table4}
\resizebox{0.85\linewidth}{!}{
\begin{tabular}{c|cc|cc}
\hline
\multirow{2}{*}{Model} & \multicolumn{2}{c|}{SJTU} & \multicolumn{2}{c}{WPC}     \\ \cline{2-5} 
                       & SRCC$\uparrow$   & PLCC$\uparrow$  & SRCC$\uparrow$ & PLCC$\uparrow$ \\ \hline
FC Layer with $P_H$           & 0.8667         & 0.8941        & 0.6819       & 0.6989       \\
FC Layer with $P_V$            & 0.8705         & 0.8889        & 0.6934       & 0.7000       \\
FC Layer with $P_{HV}$        & 0.8849         & 0.8944        & 0.7075       & 0.7126       \\
GCN with $P_{HV}$                & \bf{0.9108}    & \bf{0.9301}   & \bf{0.8054}  & \bf{0.8091}  \\ \hline
\end{tabular}}
\end{table}

\begin{table}[t]
\centering
\renewcommand\arraystretch{1.2}
\caption{Performance comparison results of different backbones on SJTU and WPC databases. The best performance is indicated in \bf{BOLD}.}  % 表格标题
\label{table5}
\resizebox{0.85\linewidth}{!}{
\begin{tabular}{c|cc|cc}
\hline
\multirow{2}{*}{Backbone} & \multicolumn{2}{c|}{SJTU} & \multicolumn{2}{c}{WPC}   \\ \cline{2-5} 
                          & SRCC$\uparrow$   & PLCC$\uparrow$  & SRCC$\uparrow$ & PLCC$\uparrow$ \\ \hline
VGG16                     & 0.8876         & 0.9148        & 0.7322       & 0.7444       \\
ResNet50                  & 0.8930         & 0.9138        & 0.7210       & 0.7288       \\
ResNet101                 & \bf{0.9108}    & \bf{0.9301}   & \bf{0.8054}  & \bf{0.8091}  \\
InceptionV3               & 0.8605         & 0.9008        & 0.6606       & 0.6688       \\
DenseNet121               & 0.8764         & 0.9124        & 0.6962       & 0.7063       \\
MobileNetV2               & 0.8829         & 0.9155        & 0.7382       & 0.7282       \\
EfficientNet-B0           & 0.8877         & 0.9094        & 0.7499       & 0.7528       \\ \hline
\end{tabular}}
\end{table}

\subsection{Ablation Study}
\subsubsection{Contributions of Multi-view Projection and GCN}
Our proposed method uses multi-view projection and GCN to improve the network performance. To measure the contributions of these operations, we conduct ablation experiments while keeping the default experimental settings unchanged. The experimental results are shown in Table \ref{table4}. Here, $P_H$ and $P_V$ represent the horizontally and vertically projected image groups, respectively. $P_{HV}$ means using both horizontally and vertically projected image groups. On the one hand, both the horizontal and vertical projection image groups contribute to the model performance, and the combination of the two can achieve better results. On the other hand, GCN can help the model better aggregate the feature information of multi-view projections, and obtain larger performance improvement on the more complex WPC database.

\subsubsection{Backbone Comparison}
In the network, 2D-CNN plays an important role as a feature extractor of images. To compare the performance of different backbones, we conduct experiments by replacing the backbones while keeping the other modules unchanged. Specifically, the backbones used for comparison include VGG16\cite{simonyan2014very}, ResNet50\cite{he2016deep}, ResNet101\cite{he2016deep}, InceptionV3\cite{szegedy2016rethink}, DenseNet\cite{huang2017dense}, MobileNetV2\cite{sandler2018mobilenetv2}, and EfficientNet\cite{tan2019efficientnet}. The experimental results are presented in Table \ref{table5}. It can be seen that the networks with other backbones for feature extraction are still competitive with state-of-the-art PCQA metrics, and there will be no significant fluctuations. Among them, ResNet101 has the best performance when used as the backbone.

\begin{table}[t]
\centering
\renewcommand\arraystretch{1.2}
\caption{Performance comparison for different number of images on SJTU and WPC databases. The best performance is indicated in \bf{BOLD}.} 
\label{table6}
\resizebox{0.85\linewidth}{!}{
\begin{tabular}{c|cc|cc}
\hline
\multirow{2}{*}{Number of Images} & \multicolumn{2}{c|}{SJTU} & \multicolumn{2}{c}{WPC}     \\ 
\cline{2-5} 
             & SRCC$\uparrow$   & PLCC$\uparrow$  & SRCC$\uparrow$ & PLCC$\uparrow$ \\ \hline
30 (RS=$24^\circ$)   & 0.8963         & 0.9150        & 0.7702       & 0.7625       \\
20 (RS=$36^\circ$)   & \bf{0.9108}    & \bf{0.9301}   & \bf{0.8054}  & \bf{0.8091}  \\
14 (RS=$48^\circ$)   & 0.8936         & 0.9159        & 0.7474       & 0.7447       \\
12 (RS=$60^\circ$)   & 0.8923         & 0.9178        & 0.7533       & 0.7430       \\ \hline
\end{tabular}}
\end{table}

\begin{figure}[t]
	\centering
	\includegraphics[width=\linewidth]{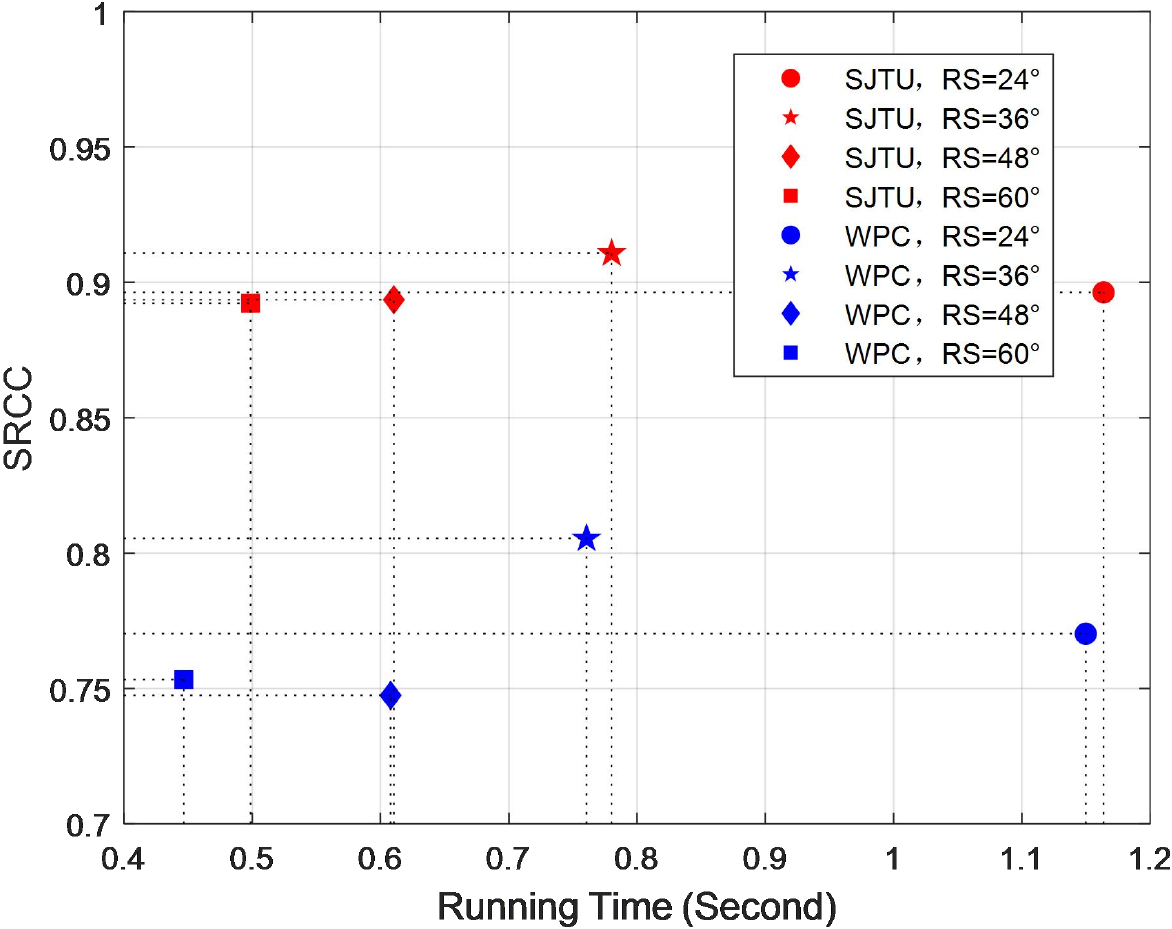}
	\caption{Running time vs. SRCC performance of GC-PCQA under four different numbers of projected images on SJTU and WPC databases. The red and blue points represent the results of the PCQA metric on the SJTU and WPC databases, respectively. RS represents the rotation stride.}
	\label{fig_6}
\end{figure}

\subsubsection{Comparison of the Number of Images}
We also conduct experiments for choosing an appropriate RS for projection to obtain a more effective multi-view projection image group $P$. In the experiments, we choose four RSs, i.e., $24^\circ$, $36^\circ$, $48^\circ$, $60^\circ$ for verification. The performance comparison results are shown in Table \ref{table6}. 
Different numbers of images are obtained by point cloud projection with different rotation strides. For example, based on the RS of $36^\circ$, the final number of multi-view projection images is $(360^\circ/36^\circ) \times 2=20$. 
Experimental results reveal that when the number of images reaches 20, the optimal performance is achieved. When the number of images becomes larger than 20, it does not lead to additional performance gain.

To compare the inference efficiency of our GC-PCQA metric under four different numbers of projected images, we report the average running time for processing a point cloud. The details are shown in Fig. \ref{fig_6}. It can be observed that although larger strides can produce fewer images and speed up network training, more projected images can bring more feature information and get better results. When the RS is reduced to $36^\circ$, the network performance reaches the limit. Further reducing the stride will negatively affect the performance. This is because the projected image at this time has been able to cover most contents of the point cloud, and more images will lead to a large number of redundancy feature information. In the optimal configuration, our method only requires 0.8 seconds, which is able to achieve real-time inference. In addition, when the RS is $60^\circ$, the network still achieves good performance while requiring only half of the inference time of the best configuration.

\begin{table}[!t]
\centering
\renewcommand\arraystretch{1.2}
\caption{Cross-database performance evaluation is performed on SJTU and WPC databases, with SJTU→WPC representing training on SJTU entire database and testing on WPC entire database. WPC→SJTU is the opposite. The best performance is indicated in \bf{BOLD}.}  % 表格标题
\label{table7}
\resizebox{0.9\linewidth}{!}{
\begin{tabular}{c|cc|cc}
\hline
\multirow{2}{*}{Method}   & \multicolumn{2}{c|}{SJTU→WPC}  & \multicolumn{2}{c|}{WPC→SJTU} \\ \cline{2-5}
          & SRCC$\uparrow$   & PLCC$\uparrow$  & SRCC$\uparrow$ & PLCC$\uparrow$   \\ \hline
3D-NSS    & 0.1352       & 0.0731        & 0.1824       & 0.1248     \\
ResSCNN   & 0.2329       & 0.2771        & 0.4031       & 0.4089     \\
PQANet    & 0.1177       & 0.3007        & 0.3141       & 0.3245     \\
IT-PCQA   & 0.1949       & 0.1966        & 0.3105       & 0.3185     \\
\bf{GC-PCQA (Ours)} & \bf{0.2404}  & \bf{0.3216}   & \bf{0.5698}  & \bf{0.6174}\\ \hline
\end{tabular}}
\end{table}

\subsection{Cross-database Evaluation}
To further verify the generalization ability, we conduct cross-database evaluation on the SJTU or WPC databases by training on one database and testing on another.
The results are shown in Table \ref{table7}. 
From the table, we can observe that the model trained on the SJTU database performs poorly on the WPC database, while the model trained on the WPC database has relatively higher performance on the SJTU database.
This is because the WPC database contains more point clouds with more complex distortions than the SJTU database. Therefore, models trained on it can achieve better performance on simpler databases. Finally, our proposed method achieves the best performance on both cases, which proves the stronger generalization capability of our model.

\section{Conclusion}
In this paper, we propose a new GCN-based NR-PCQA method called GC-PCQA. Our main inspiration comes from the fact that the HVS depends on a set of projected images from multiple viewpoints when perceiving 3D objects and the mutual dependencies among different projected images can be well modeled by GCN. Therefore, we first project the point cloud in the horizontal and vertical directions to obtain the projected image group under multiple views. Then, graph construction is performed on the projected image group and GCN is used to model the mutual dependencies between different projected 2D image contents to aggregate the feature information of images under different viewpoints, so as to imitate the viewing behavior of HVS and better measure the quality of point cloud. The experimental results on two benchmark datasets show that our proposed method has better performance than other state-of-the-art methods.

\bibliographystyle{IEEEtran}
\bibliography{reference}

\begin{thebibliography}{10}
\providecommand{\url}[1]{#1}
\csname url@samestyle\endcsname
\providecommand{\newblock}{\relax}
\providecommand{\bibinfo}[2]{#2}
\providecommand{\BIBentrySTDinterwordspacing}{\spaceskip=0pt\relax}
\providecommand{\BIBentryALTinterwordstretchfactor}{4}
\providecommand{\BIBentryALTinterwordspacing}{\spaceskip=\fontdimen2\font plus
\BIBentryALTinterwordstretchfactor\fontdimen3\font minus
  \fontdimen4\font\relax}
\providecommand{\BIBforeignlanguage}[2]{{%
\expandafter\ifx\csname l@#1\endcsname\relax
\typeout{** WARNING: IEEEtran.bst: No hyphenation pattern has been}%
\typeout{** loaded for the language `#1'. Using the pattern for}%
\typeout{** the default language instead.}%
\else
\language=\csname l@#1\endcsname
\fi
#2}}
\providecommand{\BIBdecl}{\relax}
\BIBdecl

\bibitem{liu2020model}
Q.~Liu, H.~Yuan, J.~Hou, R.~Hamzaoui, and H.~Su, ``Model-based joint bit
  allocation between geometry and color for video-based {3D} point cloud
  compression,'' \emph{IEEE Transactions on Multimedia}, vol.~23, pp.
  3278--3291, 2021.

\bibitem{huang2020pf}
Z.~Huang, Y.~Yu, J.~Xu, F.~Ni, and X.~Le, ``{PF-Net}: Point fractal network for
  {3D} point cloud completion,'' in \emph{Proceedings of the IEEE/CVF
  Conference on Computer Vision and Pattern Recognition}, 2020, pp. 7659--7667.

\bibitem{chen2020vis}
R.~Chen, S.~Han, J.~Xu, and H.~Su, ``Visibility-aware point-based multi-view
  stereo network,'' \emph{IEEE Transactions on Pattern Analysis and Machine
  Intelligence}, vol.~43, no.~10, pp. 3695--3708, 2021.

\bibitem{PointNet}
C.~R. Qi, H.~Su, K.~Mo, and L.~J. Guibas, ``Pointnet: Deep learning on point
  sets for {3D} classification and segmentation,'' in \emph{Proceedings of the
  IEEE Conference on Computer Vision and Pattern Recognition}, 2017, pp.
  652--660.

\bibitem{PointNet++}
C.~R. Qi, L.~Yi, H.~Su, and L.~J. Guibas, ``{PointNet++}: Deep hierarchical
  feature learning on point sets in a metric space,'' \emph{Advances in Neural
  Information Processing Systems}, vol.~30, 2017.

\bibitem{demisse2018def}
G.~G. Demisse, D.~Aouada, and B.~Ottersten, ``Deformation-based {3D} facial
  expression representation,'' \emph{ACM Transactions on Multimedia Computing,
  Communications, and Applications (TOMM)}, vol.~14, no.~1s, pp. 1--22, 2018.

\bibitem{javaheri2020point}
A.~Javaheri, C.~Brites, F.~Pereira, and J.~Ascenso, ``Point cloud rendering
  after coding: Impacts on subjective and objective quality,'' \emph{IEEE
  Transactions on Multimedia}, vol.~23, pp. 4049--4064, 2020.

\bibitem{cui2021deep}
Y.~Cui, R.~Chen, W.~Chu, L.~Chen, D.~Tian, Y.~Li, and D.~Cao, ``Deep learning
  for image and point cloud fusion in autonomous driving: A review,''
  \emph{IEEE Transactions on Intelligent Transportation Systems}, vol.~23,
  no.~2, pp. 722--739, 2022.

\bibitem{alexiou2020pointxr}
E.~Alexiou, N.~Yang, and T.~Ebrahimi, ``Pointxr: A toolbox for visualization
  and subjective evaluation of point clouds in virtual reality,'' in \emph{2020
  Twelfth International Conference on Quality of Multimedia Experience
  (QoMEX)}.\hskip 1em plus 0.5em minus 0.4em\relax IEEE, 2020, pp. 1--6.

\bibitem{schwarz2018mpeg}
S.~Schwarz, M.~Preda, V.~Baroncini, M.~Budagavi, P.~Cesar, P.~A. Chou, R.~A.
  Cohen, M.~Krivoku{\'c}a, S.~Lasserre, Z.~Li \emph{et~al.}, ``Emerging mpeg
  standards for point cloud compression,'' \emph{IEEE Journal on Emerging and
  Selected Topics in Circuits and Systems}, vol.~9, no.~1, pp. 133--148, 2019.

\bibitem{gu20203d}
S.~Gu, J.~Hou, H.~Zeng, and H.~Yuan, ``{3D} point cloud attribute compression
  via graph prediction,'' \emph{IEEE Signal Processing Letters}, vol.~27, pp.
  176--180, 2020.

\bibitem{vpcc}
D.~Group and et~al., ``Text of iso/iec cd 23090-5: Video-based point cloud
  compression,'' \emph{ISO/IEC JTC1/SC29/WG11 Doc. N18030}, 2018.

\bibitem{gpcc}
D.~Group \emph{et~al.}, ``Text of iso/iec cd 23090-9 geometry-based point cloud
  compression,'' \emph{ISO/IEC JTC1/SC29/WG11 Doc. N18478}, 2019.

\bibitem{liu2020coarse}
Q.~Liu, H.~Yuan, R.~Hamzaoui, and H.~Su, ``Coarse to fine rate control for
  region-based {3D} point cloud compression,'' in \emph{2020 IEEE International
  Conference on Multimedia \& Expo Workshops (ICMEW)}.\hskip 1em plus 0.5em
  minus 0.4em\relax IEEE, 2020, pp. 1--6.

\bibitem{xu2021point}
J.~Xu, Z.~Fang, Y.~Gao, S.~Ma, Y.~Jin, H.~Zhou, and A.~Wang, ``{Point
  AE-DCGAN}: A deep learning model for {3D} point cloud lossy geometry
  compression,'' in \emph{2021 Data Compression Conference (DCC)}.\hskip 1em
  plus 0.5em minus 0.4em\relax IEEE, 2021, pp. 379--379.

\bibitem{perry2020quality}
S.~Perry, H.~P. Cong, L.~A. da~Silva~Cruz, J.~Prazeres, M.~Pereira,
  A.~Pinheiro, E.~Dumic, E.~Alexiou, and T.~Ebrahimi, ``Quality evaluation of
  static point clouds encoded using mpeg codecs,'' in \emph{2020 IEEE
  International Conference on Image Processing (ICIP)}.\hskip 1em plus 0.5em
  minus 0.4em\relax IEEE, 2020, pp. 3428--3432.

\bibitem{hua2021bqe}
L.~Hua, G.~Jiang, M.~Yu, and Z.~He, ``{BQE-CVP}: Blind quality evaluator for
  colored point cloud based on visual perception,'' in \emph{2021 IEEE
  International Symposium on Broadband Multimedia Systems and Broadcasting
  (BMSB)}.\hskip 1em plus 0.5em minus 0.4em\relax IEEE, 2021, pp. 1--6.

\bibitem{zhang2022no}
Z.~Zhang, W.~Sun, X.~Min, T.~Wang, W.~Lu, and G.~Zhai, ``No-reference quality
  assessment for {3D} colored point cloud and mesh models,'' \emph{IEEE
  Transactions on Circuits and Systems for Video Technology}, vol.~32, no.~11,
  pp. 7618--7631, 2022.

\bibitem{chetouani2021deep}
A.~Chetouani, M.~Quach, G.~Valenzise, and F.~Dufaux, ``Deep learning-based
  quality assessment of {3D} point clouds without reference,'' in \emph{2021
  IEEE International Conference on Multimedia \& Expo Workshops (ICMEW)}.\hskip
  1em plus 0.5em minus 0.4em\relax IEEE, 2021, pp. 1--6.

\bibitem{liu2023point}
Y.~Liu, Q.~Yang, Y.~Xu, and L.~Yang, ``Point cloud quality assessment: Dataset
  construction and learning-based no-reference metric,'' \emph{ACM Transactions
  on Multimedia Computing, Communications and Applications}, vol.~19, no.~2s,
  pp. 1--26, 2023.

\bibitem{tao2021point}
W.-x. Tao, G.-y. Jiang, Z.-d. Jiang, and M.~Yu, ``Point cloud projection and
  multi-scale feature fusion network based blind quality assessment for colored
  point clouds,'' in \emph{Proceedings of the 29th ACM International Conference
  on Multimedia}, 2021, pp. 5266--5272.

\bibitem{liu2021pqa}
Q.~Liu, H.~Yuan, H.~Su, H.~Liu, Y.~Wang, H.~Yang, and J.~Hou, ``{PQA-Net}: Deep
  no reference point cloud quality assessment via multi-view projection,''
  \emph{IEEE Transactions on Circuits and Systems for Video Technology},
  vol.~31, no.~12, pp. 4645--4660, 2021.

\bibitem{tu2022v}
R.~Tu, G.~Jiang, M.~Yu, T.~Luo, Z.~Peng, and F.~Chen, ``{V-PCC} projection
  based blind point cloud quality assessment for compression distortion,''
  \emph{IEEE Transactions on Emerging Topics in Computational Intelligence},
  vol.~7, no.~2, pp. 462--473, 2023.

\bibitem{yang2022no}
Q.~Yang, Y.~Liu, S.~Chen, Y.~Xu, and J.~Sun, ``No-reference point cloud quality
  assessment via domain adaptation,'' in \emph{Proceedings of the IEEE/CVF
  Conference on Computer Vision and Pattern Recognition}, June 2022, pp.
  21\,179--21\,188.

\bibitem{wu2020compre}
Z.~Wu, S.~Pan, F.~Chen, G.~Long, C.~Zhang, and S.~Y. Philip, ``A comprehensive
  survey on graph neural networks,'' \emph{IEEE Transactions on Neural Networks
  and Learning Systems}, vol.~32, no.~1, pp. 4--24, 2021.

\bibitem{kipf2016semi}
T.~N. Kipf and M.~Welling, ``Semi-supervised classification with graph
  convolutional networks,'' \emph{arXiv preprint arXiv:1609.02907}, 2016.

\bibitem{zhang2018graph}
Y.~Zhang and M.~Rabbat, ``A graph-{CNN} for {3D} point cloud classification,''
  in \emph{2018 IEEE International Conference on Acoustics, Speech and Signal
  Processing (ICASSP)}.\hskip 1em plus 0.5em minus 0.4em\relax IEEE, 2018, pp.
  6279--6283.

\bibitem{li2021structural}
Y.~Li and Y.~Tanaka, ``Structural features in feature space for structure-aware
  graph convolution,'' in \emph{2021 IEEE International Conference on Image
  Processing (ICIP)}.\hskip 1em plus 0.5em minus 0.4em\relax IEEE, 2021, pp.
  3158--3162.

\bibitem{xu2019msgcnn}
M.~Xu, W.~Dai, Y.~Shen, and H.~Xiong, ``{MSGCNN}: Multi-scale graph
  convolutional neural network for point cloud segmentation,'' in \emph{2019
  IEEE Fifth International Conference on Multimedia Big Data (BigMM)}.\hskip
  1em plus 0.5em minus 0.4em\relax IEEE, 2019, pp. 118--127.

\bibitem{hansen2018multi}
L.~Hansen, J.~Diesel, and M.~P. Heinrich, ``Multi-kernel diffusion {CNN}s for
  graph-based learning on point clouds,'' in \emph{Proceedings of the European
  Conference on Computer Vision (ECCV) Workshops}, September 2018.

\bibitem{si2019action}
C.~Si, W.~Chen, W.~Wang, L.~Wang, and T.~Tan, ``An attention enhanced graph
  convolutional {LSTM} network for skeleton-based action recognition,'' in
  \emph{Proceedings of the IEEE/CVF Conference on Computer Vision and Pattern
  Recognition}, June 2019, pp. 1227--1236.

\bibitem{sun2022graphiqa}
S.~Sun, T.~Yu, J.~Xu, W.~Zhou, and Z.~Chen, ``{GraphIQA}: Learning distortion
  graph representations for blind image quality assessment,'' \emph{IEEE
  Transactions on Multimedia}, 2022.

\bibitem{huang2022explainable}
Y.~Huang, L.~Li, Y.~Yang, Y.~Li, and Y.~Guo, ``Explainable and generalizable
  blind image quality assessment via semantic attribute reasoning,'' \emph{IEEE
  Transactions on Multimedia}, 2022.

\bibitem{xu2020blind}
J.~Xu, W.~Zhou, and Z.~Chen, ``Blind omnidirectional image quality assessment
  with viewport oriented graph convolutional networks,'' \emph{IEEE
  Transactions on Circuits and Systems for Video Technology}, vol.~31, no.~5,
  pp. 1724--1737, 2020.

\bibitem{fu2022adaptive}
J.~Fu, C.~Hou, W.~Zhou, J.~Xu, and Z.~Chen, ``Adaptive hypergraph convolutional
  network for no-reference 360-degree image quality assessment,'' in
  \emph{Proceedings of the 30th ACM International Conference on Multimedia},
  2022, pp. 961--969.

\bibitem{abouelaziz2021learning}
I.~Abouelaziz, A.~Chetouani, M.~El~Hassouni, H.~Cherifi, and L.~J. Latecki,
  ``Learning graph convolutional network for blind mesh visual quality
  assessment,'' \emph{IEEE Access}, vol.~9, pp. 108\,200--108\,211, 2021.

\bibitem{el2022learning}
M.~El~Hassouni and H.~Cherifi, ``Learning graph features for colored mesh
  visual quality assessment,'' in \emph{2022 IEEE International Conference on
  Image Processing (ICIP)}.\hskip 1em plus 0.5em minus 0.4em\relax IEEE, 2022,
  pp. 3381--3385.

\bibitem{mekuria2016eval}
R.~Mekuria, Z.~Li, C.~Tulvan, and P.~Chou, ``Evaluation criteria for point
  cloud compression,'' \emph{ISO/IEC MPEG}, no. 16332, 2016.

\bibitem{tian2017geometric}
D.~Tian, H.~Ochimizu, C.~Feng, R.~Cohen, and A.~Vetro, ``Geometric distortion
  metrics for point cloud compression,'' in \emph{2017 IEEE International
  Conference on Image Processing (ICIP)}.\hskip 1em plus 0.5em minus
  0.4em\relax IEEE, 2017, pp. 3460--3464.

\bibitem{alexiou2018point}
E.~Alexiou and T.~Ebrahimi, ``Point cloud quality assessment metric based on
  angular similarity,'' in \emph{2018 IEEE International Conference on
  Multimedia and Expo (ICME)}.\hskip 1em plus 0.5em minus 0.4em\relax IEEE,
  2018, pp. 1--6.

\bibitem{javaheri2020gen}
A.~Javaheri, C.~Brites, F.~Pereira, and J.~Ascenso, ``A generalized hausdorff
  distance based quality metric for point cloud geometry,'' in \emph{2020
  Twelfth International Conference on Quality of Multimedia Experience
  (QoMEX)}.\hskip 1em plus 0.5em minus 0.4em\relax IEEE, 2020, pp. 1--6.

\bibitem{mekuria2017per}
R.~Mekuria, S.~Laserre, and C.~Tulvan, ``Performance assessment of point cloud
  compression,'' in \emph{2017 IEEE Visual Communications and Image Processing
  (VCIP)}.\hskip 1em plus 0.5em minus 0.4em\relax IEEE, 2017, pp. 1--4.

\bibitem{viola2020color}
I.~Viola, S.~Subramanyam, and P.~Cesar, ``A color-based objective quality
  metric for point cloud contents,'' in \emph{2020 Twelfth International
  Conference on Quality of Multimedia Experience (QoMEX)}.\hskip 1em plus 0.5em
  minus 0.4em\relax IEEE, 2020, pp. 1--6.

\bibitem{meynet2019pc}
G.~Meynet, J.~Digne, and G.~Lavou{\'e}, ``{PC-MSDM}: A quality metric for {3D}
  point clouds,'' in \emph{2019 Eleventh International Conference on Quality of
  Multimedia Experience (QoMEX)}.\hskip 1em plus 0.5em minus 0.4em\relax IEEE,
  2019, pp. 1--3.

\bibitem{meynet2020pcqm}
G.~Meynet, Y.~Nehm{\'e}, J.~Digne, and G.~Lavou{\'e}, ``{PCQM}: A
  full-reference quality metric for colored {3D} point clouds,'' in \emph{2020
  Twelfth International Conference on Quality of Multimedia Experience
  (QoMEX)}.\hskip 1em plus 0.5em minus 0.4em\relax IEEE, 2020, pp. 1--6.

\bibitem{alexiou2020to}
E.~Alexiou and T.~Ebrahimi, ``Towards a point cloud structural similarity
  metric,'' in \emph{2020 IEEE International Conference on Multimedia \& Expo
  Workshops (ICMEW)}.\hskip 1em plus 0.5em minus 0.4em\relax IEEE, 2020, pp.
  1--6.

\bibitem{diniz2020to}
R.~Diniz, P.~G. Freitas, and M.~C. Farias, ``Towards a point cloud quality
  assessment model using local binary patterns,'' in \emph{2020 Twelfth
  International Conference on Quality of Multimedia Experience (QoMEX)}.\hskip
  1em plus 0.5em minus 0.4em\relax IEEE, 2020, pp. 1--6.

\bibitem{diniz2020multi}
R.~Diniz, P.~G. Freitas, and et~al., ``Multi-distance point cloud quality
  assessment,'' in \emph{2020 IEEE International Conference on Image Processing
  (ICIP)}.\hskip 1em plus 0.5em minus 0.4em\relax IEEE, 2020, pp. 3443--3447.

\bibitem{diniz2020local}
R.~Diniz, P.~G. Freitas, and M.~C. Farias, ``Local luminance patterns for point
  cloud quality assessment,'' in \emph{2020 IEEE 22nd International Workshop on
  Multimedia Signal Processing (MMSP)}.\hskip 1em plus 0.5em minus 0.4em\relax
  IEEE, 2020, pp. 1--6.

\bibitem{yang2020infer}
Q.~Yang, Z.~Ma, Y.~Xu, Z.~Li, and J.~Sun, ``Inferring point cloud quality via
  graph similarity,'' \emph{IEEE Transactions on Pattern Analysis and Machine
  Intelligence}, vol.~44, no.~6, pp. 3015--3029, 2022.

\bibitem{yang2022mped}
Q.~Yang, Y.~Zhang, S.~Chen, Y.~Xu, J.~Sun, and Z.~Ma, ``{MPED}: Quantifying
  point cloud distortion based on multiscale potential energy discrepancy,''
  \emph{IEEE Transactions on Pattern Analysis and Machine Intelligence},
  vol.~45, no.~5, pp. 6037--6054, 2023.

\bibitem{viola2020red}
I.~Viola and P.~Cesar, ``A reduced reference metric for visual quality
  evaluation of point cloud contents,'' \emph{IEEE Signal Processing Letters},
  vol.~27, pp. 1660--1664, 2020.

\bibitem{liu2021reduced}
Q.~Liu, H.~Yuan, R.~Hamzaoui, H.~Su, J.~Hou, and H.~Yang, ``Reduced reference
  perceptual quality model with application to rate control for video-based
  point cloud compression,'' \emph{IEEE Transactions on Image Processing},
  vol.~30, pp. 6623--6636, 2021.

\bibitem{zhou2022blind}
W.~Zhou, Q.~Yang, Q.~Jiang, G.~Zhai, and W.~Lin, ``Blind quality assessment of
  {3D} dense point clouds with structure guided resampling,'' \emph{arXiv
  preprint arXiv:2208.14603}, 2022.

\bibitem{wolf2009ref}
S.~Wolf and M.~Pinson, ``Reference algorithm for computing peak signal to noise
  ratio (psnr) of a video sequence with a constant delay,'' \emph{ITU-T
  Contribution COM9-C6-E}, 2009.

\bibitem{wang2004image}
Z.~Wang, A.~C. Bovik, H.~R. Sheikh, and E.~P. Simoncelli, ``Image quality
  assessment: from error visibility to structural similarity,'' \emph{IEEE
  Transactions on Image Processing}, vol.~13, no.~4, pp. 600--612, 2004.

\bibitem{wang2003multi}
Z.~Wang, E.~P. Simoncelli, and A.~C. Bovik, ``Multiscale structural similarity
  for image quality assessment,'' in \emph{The Thrity-Seventh Asilomar
  Conference on Signals, Systems \& Computers, 2003}, vol.~2.\hskip 1em plus
  0.5em minus 0.4em\relax Ieee, 2003, pp. 1398--1402.

\bibitem{wang2010info}
Z.~Wang and Q.~Li, ``Information content weighting for perceptual image quality
  assessment,'' \emph{IEEE Transactions on Image Processing}, vol.~20, no.~5,
  pp. 1185--1198, 2011.

\bibitem{sheikh2006image}
H.~R. Sheikh and A.~C. Bovik, ``Image information and visual quality,''
  \emph{IEEE Transactions on Image Processing}, vol.~15, no.~2, pp. 430--444,
  2006.

\bibitem{freitas2022point}
X.~G. Freitas, R.~Diniz, and M.~C. Farias, ``Point cloud quality assessment:
  unifying projection, geometry, and texture similarity,'' \emph{The Visual
  Computer}, pp. 1--8, 2022.

\bibitem{chen2017contour}
S.~Chen, D.~Tian, C.~Feng, A.~Vetro, and J.~Kova{\v{c}}evi{\'c},
  ``Contour-enhanced resampling of {3D} point clouds via graphs,'' in
  \emph{2017 IEEE International Conference on Acoustics, Speech and Signal
  Processing (ICASSP)}.\hskip 1em plus 0.5em minus 0.4em\relax IEEE, 2017, pp.
  2941--2945.

\bibitem{chen2017fast}
S.~Chen, D.~Tian, C.~Feng, A.~Vetro, and et~al., ``Fast resampling of
  three-dimensional point clouds via graphs,'' \emph{IEEE Transactions on
  Signal Processing}, vol.~66, no.~3, pp. 666--681, 2018.

\bibitem{qi2019feature}
J.~Qi, W.~Hu, and Z.~Guo, ``Feature preserving and uniformity-controllable
  point cloud simplification on graph,'' in \emph{2019 IEEE International
  Conference on Multimedia and Expo (ICME)}.\hskip 1em plus 0.5em minus
  0.4em\relax IEEE, 2019, pp. 284--289.

\bibitem{he2016deep}
K.~He, X.~Zhang, S.~Ren, and J.~Sun, ``Deep residual learning for image
  recognition,'' in \emph{Proceedings of the IEEE Conference on Computer Vision
  and Pattern Recognition}, June 2016, pp. 770--778.

\bibitem{hu2018se}
J.~Hu, L.~Shen, and G.~Sun, ``Squeeze-and-excitation networks,'' in
  \emph{Proceedings of the IEEE Conference on Computer Vision and Pattern
  Recognition}, June 2018, pp. 7132--7141.

\bibitem{woo2018cbam}
S.~Woo, J.~Park, J.-Y. Lee, and I.~S. Kweon, ``{CBAM}: Convolutional block
  attention module,'' in \emph{Proceedings of the European Conference on
  Computer Vision (ECCV)}, September 2018, pp. 3--19.

\bibitem{roy2018concurrent}
A.~G. Roy, N.~Navab, and C.~Wachinger, ``Concurrent spatial and channel
  ‘squeeze \& excitation’in fully convolutional networks,'' in
  \emph{Medical Image Computing and Computer Assisted Intervention--MICCAI
  2018: 21st International Conference, Granada, Spain, September 16-20, 2018,
  Proceedings, Part I}.\hskip 1em plus 0.5em minus 0.4em\relax Springer, 2018,
  pp. 421--429.

\bibitem{chaib2017deep}
S.~Chaib, H.~Liu, Y.~Gu, and H.~Yao, ``Deep feature fusion for vhr remote
  sensing scene classification,'' \emph{IEEE Transactions on Geoscience and
  Remote Sensing}, vol.~55, no.~8, pp. 4775--4784, 2017.

\bibitem{akilan2017late}
T.~Akilan, Q.~J. Wu, A.~Safaei, and W.~Jiang, ``A late fusion approach for
  harnessing multi-cnn model high-level features,'' in \emph{2017 IEEE
  International Conference on Systems, Man, and Cybernetics (SMC)}.\hskip 1em
  plus 0.5em minus 0.4em\relax IEEE, 2017, pp. 566--571.

\bibitem{yan2020blind}
J.~Yan, Y.~Fang, L.~Huang, X.~Min, Y.~Yao, and G.~Zhai, ``Blind stereoscopic
  image quality assessment by deep neural network of multi-level feature
  fusion,'' in \emph{2020 IEEE International Conference on Multimedia and Expo
  (ICME)}.\hskip 1em plus 0.5em minus 0.4em\relax IEEE, 2020, pp. 1--6.

\bibitem{wu2019spatial}
C.~Wu, X.-J. Wu, and J.~Kittler, ``Spatial residual layer and dense connection
  block enhanced spatial temporal graph convolutional network for
  skeleton-based action recognition,'' in \emph{Proceedings of the IEEE/CVF
  International Conference on Computer Vision Workshops}, Oct 2019.

\bibitem{yang2020predict}
Q.~Yang, H.~Chen, Z.~Ma, Y.~Xu, R.~Tang, and J.~Sun, ``Predicting the
  perceptual quality of point cloud: A {3D-to-2D} projection-based
  exploration,'' \emph{IEEE Transactions on Multimedia}, vol.~23, pp.
  3877--3891, 2021.

\bibitem{su2019perceptual}
H.~Su, Z.~Duanmu, W.~Liu, Q.~Liu, and Z.~Wang, ``Perceptual quality assessment
  of {3D} point clouds,'' in \emph{2019 IEEE International Conference on Image
  Processing (ICIP)}.\hskip 1em plus 0.5em minus 0.4em\relax IEEE, 2019, pp.
  3182--3186.

\bibitem{video2003final}
V.~Q.~E. Group \emph{et~al.}, ``Final report from the video quality experts
  group on the validation of objective models of video quality assessment,
  phase ii,'' \emph{2003 VQEG}, 2003.

\bibitem{kingma2014adam}
D.~P. Kingma and J.~Ba, ``{Adam}: A method for stochastic optimization,''
  \emph{arXiv preprint arXiv:1412.6980}, 2014.

\bibitem{zhou2023reduced}
W.~Zhou, G.~Yue, R.~Zhang, Y.~Qin, and H.~Liu, ``Reduced-reference quality
  assessment of point clouds via content-oriented saliency projection,''
  \emph{arXiv preprint arXiv:2301.07681}, 2023.

\bibitem{simonyan2014very}
K.~Simonyan and A.~Zisserman, ``Very deep convolutional networks for
  large-scale image recognition,'' \emph{arXiv preprint arXiv:1409.1556}, 2014.

\bibitem{szegedy2016rethink}
C.~Szegedy, V.~Vanhoucke, S.~Ioffe, J.~Shlens, and Z.~Wojna, ``Rethinking the
  inception architecture for computer vision,'' in \emph{Proceedings of the
  IEEE Conference on Computer Vision and Pattern Recognition}, June 2016, pp.
  2818--2826.

\bibitem{huang2017dense}
G.~Huang, Z.~Liu, L.~Van Der~Maaten, and K.~Q. Weinberger, ``Densely connected
  convolutional networks,'' in \emph{Proceedings of the IEEE Conference on
  Computer Vision and Pattern Recognition}, July 2017, pp. 4700--4708.

\bibitem{sandler2018mobilenetv2}
M.~Sandler, A.~Howard, M.~Zhu, A.~Zhmoginov, and L.-C. Chen, ``{Mobilenetv2}:
  Inverted residuals and linear bottlenecks,'' in \emph{Proceedings of the IEEE
  Conference on Computer Vision and Pattern Recognition}, June 2018, pp.
  4510--4520.

\bibitem{tan2019efficientnet}
M.~Tan and Q.~Le, ``{EfficientNet}: Rethinking model scaling for convolutional
  neural networks,'' in \emph{International Conference on Machine Learning},
  vol.~97.\hskip 1em plus 0.5em minus 0.4em\relax PMLR, 09--15 Jun 2019, pp.
  6105--6114.

\end{thebibliography}

\vspace{-1cm}
\begin{IEEEbiography}[{\includegraphics[width=1in,height=1.33in,clip]{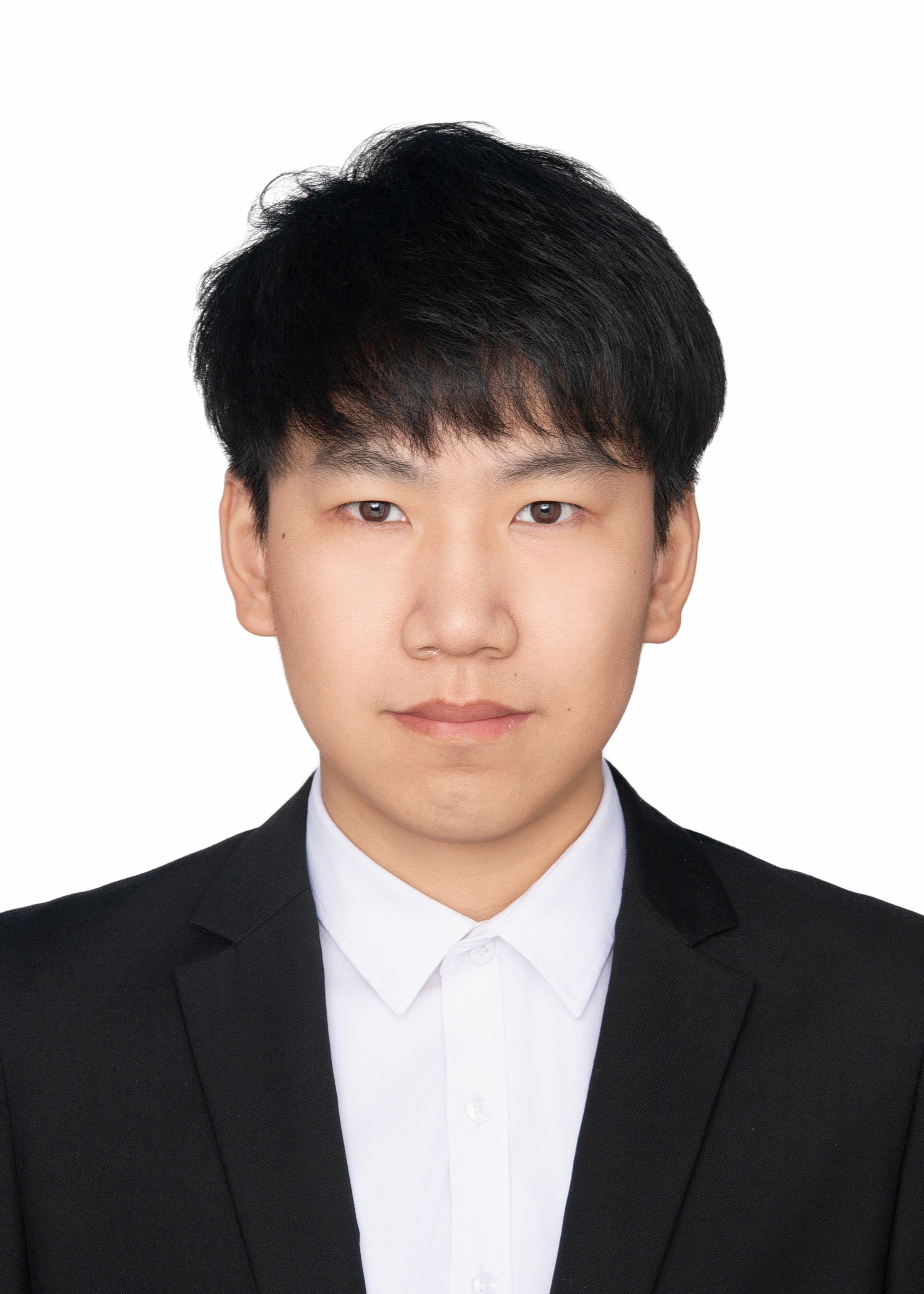}}]
{Wu Chen} received the B.E. degree in software engineering from the Hangzhou Normal University in 2022. He is currently pursuing the M.E. degree with the School of Information Science and Engineering, Ningbo University, Ningbo, China. His research interests include the application of 3D point clouds in computer vision, such as quality assessment and denoising.
\end{IEEEbiography}

\vspace{-1cm}
\begin{IEEEbiography}[{\includegraphics[width=1in,height=1.33in,clip]{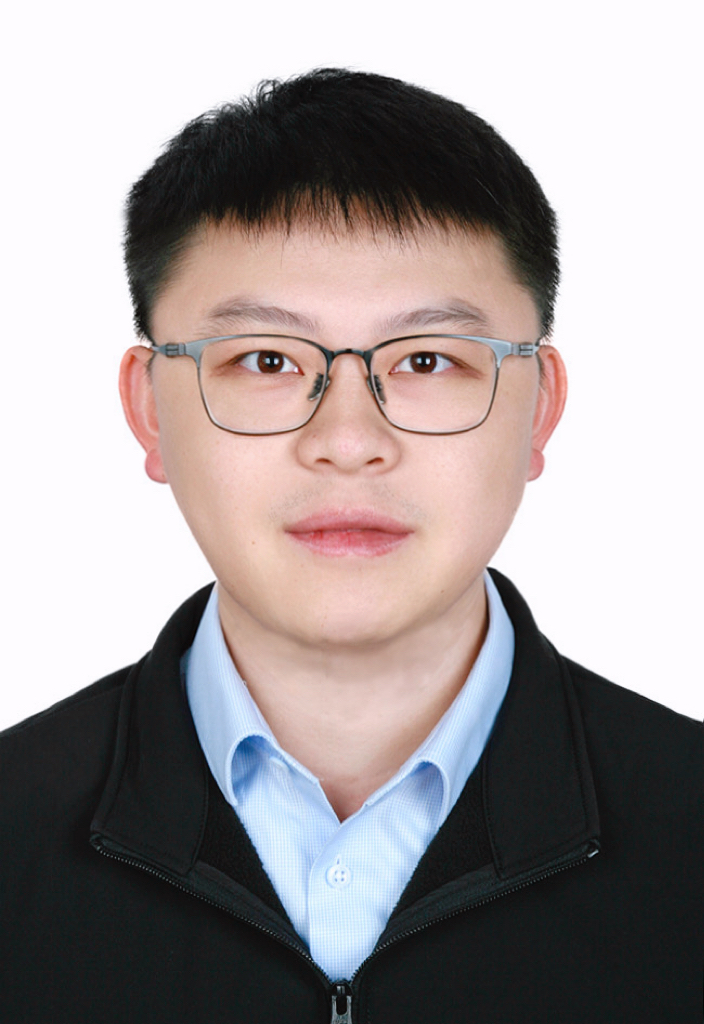}}]
{Qiuping Jiang} (Senior Member, IEEE) is currently a Professor with the School of Information Science and Engineering, Ningbo University, Ningbo, China. His research interests include image quality assessment, visual perception modeling, and water-related vision. He is an Associate Editor or on the Editorial Board of several SCI-indexed journals such as \textit{Displays}, \textit{Journal of Visual Communication and Image Representation}, \textit{IET Image Processing}, and \textit{Journal of Electronic Imaging}. 
\end{IEEEbiography}

\vspace{-1cm}
\begin{IEEEbiography}[{\includegraphics[width=1in,height=1.33in,clip]{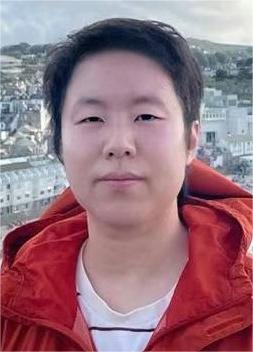}}]
{Wei Zhou} (Senior Member, IEEE) is an Assistant Professor at Cardiff University, United Kingdom. Dr. Zhou was a Postdoctoral Fellow at University of Waterloo, Canada. Wei received the Ph.D. degree from the University of Science and Technology of China in 2021, joint with the University of Waterloo from 2019 to 2021. Dr. Zhou was a visiting scholar at National Institute of Informatics, Japan, a research assistant with Intel, and a research intern at Microsoft Research and Alibaba Cloud. Wei is now an Associate Editor of IEEE Transactions on Neural Networks and Learning Systems. Wei’s research interests span multimedia computing, perceptual image processing, and computational vision.
\end{IEEEbiography}

\vspace{-1cm}
\begin{IEEEbiography}[{\includegraphics[width=1in,height=1.33in,clip]{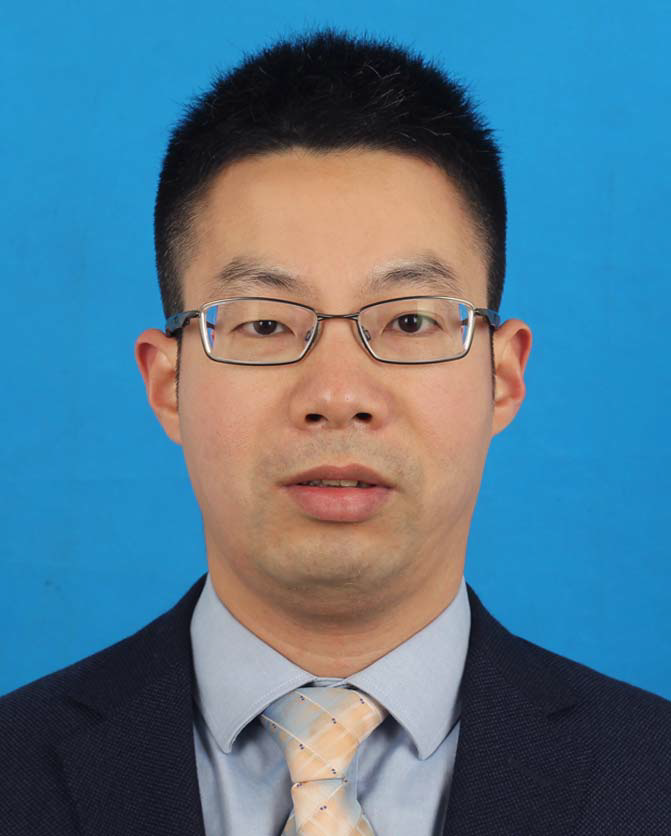}}]
{Feng Shao} (Senior Member, IEEE) received the B.S. and Ph.D. degrees in electronic science and technology from Zhejiang University, Hangzhou, China, in 2002 and 2007, respectively. He is currently a Professor with the Faculty of Information Science and Engineering, Ningbo University, China. He was a Visiting Fellow with the School of Computer Engineering, Nanyang Technological University, Singapore, from February 2012 to August 2012. He has published over 100 technical articles in refereed journals and proceedings in the areas of image processing, image quality assessment, and immersive media computing. He received the Excellent Young Scholar Award by NSF of China (NSFC) in 2016.
\end{IEEEbiography}

\vspace{-1cm}
\begin{IEEEbiography}[{\includegraphics[width=1in,height=1.33in,clip]{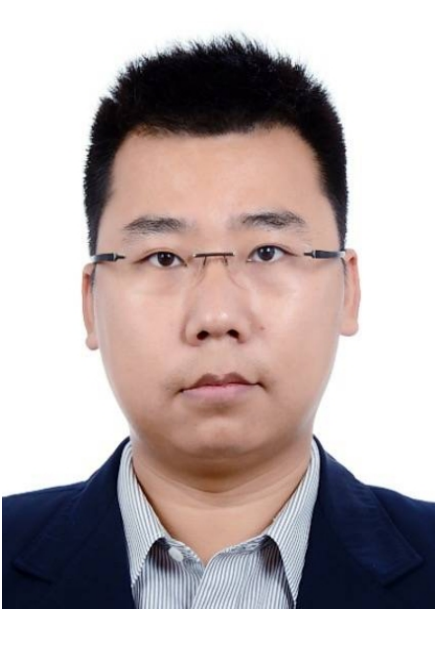}}]
{Guangtao Zhai} (Senior Member, IEEE) received the B.E. and M.E. degrees from Shandong University, Shandong, China, in 2001 and 2004, respectively, and the Ph.D. degree from Shanghai Jiao Tong University, Shanghai, China, in 2009. He is currently a Research Professor with the Institute of Image Communication and Information Processing, Shanghai Jiao Tong University. From 2008 to 2009, he was a Visiting Student with the Department of Electrical and Computer Engineering, McMaster University, Hamilton, ON, Canada, where he was a Postdoctoral Fellow from 2010 to 2012. From 2012 to 2013, he was a Humboldt Research Fellow with the Institute of Multimedia Communication and Signal Processing, Friedrich Alexander University of Erlangen-Nuremberg, Erlangen, Germany. His research interests include multimedia signal processing and perceptual signal processing. He was the recipient of the Award of National Excellent Ph.D. Thesis from the Ministry of Education of China in 2012.
\end{IEEEbiography}

\vspace{-1cm}
\begin{IEEEbiography}[{\includegraphics[width=1in,height=1.33in,clip]{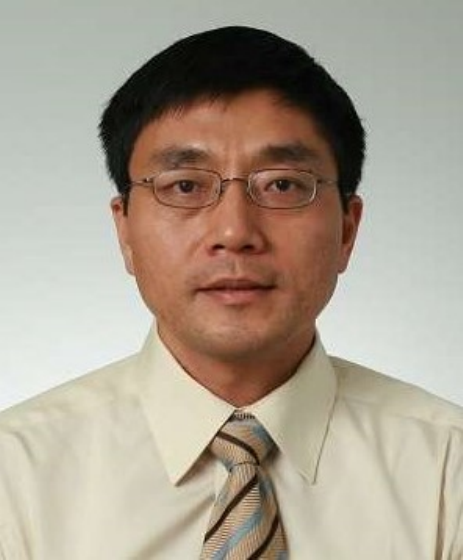}}]{Weisi Lin} (Fellow, IEEE) received the bachelor’s degree in electronics and the master’s degree in digital signal processing from Sun Yat-Sen University, Guangzhou, China, and the Ph.D. degree in computer vision from King’s College London, U.K.

He is currently a Professor with the School of Computer Science and Engineering, Nanyang Technological University, Singapore. His research interests include image processing, perceptual modeling, video compression, multimedia communication, and computer vision. He is a fellow of the IET, an Honorary Fellow of the Singapore Institute of Engineering Technologists, and a Chartered Engineer in U.K. He was awarded as a Distinguished Lecturer of the IEEE Circuits and Systems Society from 2016 to 2017. He served as a Lead Guest Editor for a Special Issue on Perceptual Signal Processing of the IEEE JSTSP in 2012. He has also served or serves as an Associate Editor for IEEE TIP, IEEE TCSVT, IEEE TMM, IEEE TNNLS, IEEE SPL, and Journal of Visual Communication and Image Representation. He was the Chair of the IEEE MMTC Special Interest Group on Quality of Experience.
\end{IEEEbiography}

\end{document}